# Adaptive Differential Evolution with Diversification: Addressing Optimization Challenges

Sarit Maitra
Alliance University, Bengaluru, India
sarit.maitra@gmail.com

**Abstract**

The existing variants of the Differential Evolution (DE) algorithm come with certain limitations, such as poor local search and susceptibility to premature convergence. This study introduces Adaptive Differential Evolution with Diversification (ADED), a method that dynamically modifies the neighborhood structure by evaluating the trial solutions' fitness. Developed to work with both convex and nonconvex objective functions, ADED is validated with 22 benchmark functions, including Rosenbrock, Rastrigin, Ackley, and DeVilliers-Glasser02. The development is carried out in Google Cloud using Jupyter Notebook and Python v3.10.12, with additional testing conducted on the multi-objective benchmark ZDT test suite. ADED distinguishes itself with its adaptive and diverse approach, which includes adaptive mutation and crossover-rates, diverse mutation tactics, diversification measurements, local search mechanisms, and convergence monitoring. The unique combination of these features collectively enhances ADED's effectiveness in navigating complex and diverse landscapes, positioning it as a promising tool for addressing challenges in both single- and multi-objective optimization scenarios.

## 1. Introduction

Optimization is a method for making decisions in complex situations involving nonlinear, multimodal, discontinuous, and non-differentiable models. Its goal is to maximize or minimize objective functions within a reasonable timeframe. While it is a well-known problem in applied machine learning and mathematical literature (Momin & Yang, 2013), it has a fair share in various business domains, including the supply chain. Optimization challenges in the supply chain can be complex and involve a variety of constraints. These challenges often necessitate the use of good optimization strategies to minimize or maximize objective functions. However, when these functions are not linear or polynomial and must be approximated, the optimization methods are mostly incapable of solving the problems. Though the use of whole or partial derivatives to linearize these functions may work in certain cases, the use of metaheuristic evolutionary algorithms for approximation has become popular among the practitioners (Abdolrasol et al., 2021). These algorithms use their search mechanisms, which draw inspiration from various natural occurrences (Ahmad MF et al., 2022). Empirical evidence suggests that metaheuristic optimization algorithms increase sustainability in supply chain while increasing efficiency and competitiveness (e.g., Abualigah et al., 2023). These algorithms have grown in popularity over time because they execute global searches while avoiding local optima without requiring explicit knowledge of the gradient or differentiability of the functions involved in the optimization problem (Ahmad et al., 2023). Metaheuristics are especially helpful for optimization problems with a lot of variables or constraints (Cuong-Le et al., 2021; Sang-To et al., 2023).

The network architecture of supply chains in the past was primarily concerned with the location of warehouses and distribution fleets. These days, maintaining a brand, getting a competitive edge, and providing real-time decision assistance all depend on optimization. However, stochastic demand patterns pose unique challenges to supply chain optimization, especially in the wake of the epidemic, which brought with it unusual uncertainty. An empirical study shows that from January 2008 to October 2020, stochastic approaches were used at 27.68%, emerging as the leading strategy for supply chain optimization (Oliveira & Machado, 2021). Techniques for handling stochastic demands become paramount in ensuring supply chain resilience and adaptability (Lan 2020; Zakaria et al., 2020). This has generated great amount of interest within the research community regarding how optimization issues can be handled more effectively during demand variation (Fallahi et al., 2022; Niu et al., 2021; Sethanan & Jamrus, 2020).

Metaheuristic algorithms do not guarantee optimal solutions compared to traditional optimization. Instead, great solutions which are close to the ideal option in a fair amount of time can be obtained using metaheuristic algorithms (Ezugwu et al., 2021). The evolutionary aspect of metaheuristic algorithms is the focus of this work. Evolutionary algorithms use a variety of genetic processes, including crossover-rate (CR), mutation, and population size, to produce new, better offspring solutions. These algorithms' exploratory properties make them valuable not only for complex supply chain management and inventory optimization, but also for a range of other disciplines. Of the metaheuristic algorithms, the Differential Evolution (DE) algorithm has become a popular evolutionary algorithm since its inception (Storn, 1996; Storn & Price, 1997). The goal was initially to solve Chebyshev polynomial problems, which eventually evolved into an effective computing technique for handling complex optimization issues. In the field of supply chain optimization, where objective functions are frequently complex and include simulations or external software, the DE algorithm proved to be quite effective (Wang et al., 2022; Guo et al., 2023; Moonsri et al., 2022 etc.). The advantage with DE algorithm is that it excels at navigating complex solution spaces and identifying optimal solutions without requiring a thorough understanding of the underlying processes. Application of DE can be found in a wide spectrum of supply chain functions (e.g., Jauhar et al., 2017; Doolun et al., 2018; Yousefi & Tosarkani, 2022; Nimmy et al., 2022; Guo et al., 2023, etc.).

However, researchers found that, DE's local search capabilities are often insufficient, making it prone to premature or poor convergence (Rauf et al., 2021) and getting trapped in local optima (Wang et al., 2022; Kononova et al., 2021; Liu et al., 2020). Configuring techniques and settings via the trial-and-error method is often time-consuming and computationally expensive (Chakraborty et al., 2023). Additionally, when applied to high-dimensional optimization problems, DE's optimization accuracy tends to diminish (Cai et al., 2019; Deng et al., 2022; Liu et al., 2023). A significant amount of research has already gone into improving classic DE algorithm (Opara & Arabas, 2019), which include DE variants where the classic DE framework has been integrated with the dynamically modifying population size and control parameters (e.g., Maitra et al., 2023; Qiao et al., 2022; Salgotra et al., 2021; Rauf et al., 2021; Deng et al., 2021; Tan et al., 2021; Chakraborty, 2008; Mallipeddi, 2011; Draa et al., 2015; Deng et al., 2021; Xu et al., 2021; Wang et al., 2022; Hadikhani et al., 2023). Pant et al. (2020) conducted a comprehensive review of over 25 years of DE research, revealing continuous adaptations and enhancements aimed at improving DE algorithm's performance on challenging optimization problems since its inception. They found that DE has gone through multiple variants such as population generation, mutation schemes, crossover schemes, variation in parameters and hybridized variants along with various successful applications of DE. The advancement in DE variants highlights the ongoing research efforts to enhance and broaden the capabilities of DE and associated optimization methods. Even with the improvement, considering its popularity, there is a growing need for in-depth research addressing these limitations and seeking potential enhancements to further elevate DE's effectiveness as an optimization tool.

Therefore, the existing limitations of the classic DE algorithm are the motivation for this study. Moreover, how to accelerate the convergence of an algorithm remains a difficult problem, thus new optimization approaches must be developed on a regular basis to progress the field of computational intelligence or heuristic optimization (Sarker et al., 2019; Xue et al., 2019). With the introduction of Adaptive Differential Evolution and Diversification, or ADED, this work aims to enhance optimization performance and address challenges associated with supply chain analytics. ADED's key features include dynamic parameter adaptation, individual-specific adaptation, crowding selection, dynamic neighborhood structures, and a simple initialization strategy. Some key highlights are that ADED adjusts the dynamic parameters (differential weight and crossover-rate to respond to the evolving optimization landscape. ADED also adds the individual-specific adaptation which allows the algorithm to tailor its exploration-exploitation strategy to diverse behaviors. ADED also incorporates crowding selection, which involves trial solutions competing against the current population based on their fitness, enhancing exploration of the solution space. Moreover, dynamic neighborhoods allow potential solutions to have different sets of neighbors, enhancing adaptability to different regions of the solution space and subsequently it adds update neighborhood. Section 2 discusses the details of the changes, followed by several benchmark test functions in Section 3 and empirical investigation and discussions in Section 4.

## 2. ADED - Adaptive Differential Evolution with Diversification

The ADED algorithm addresses the issues of premature convergence and lack of diversity in Differential Evolution variants. It falls under the category of population-based stochastic optimization methods. The primary goal is to discover the best solution for both quadratic convex and sinusoidal non-convex functions. Fig. 1 displays the landscape for the non-convex objective function, which is difficult for the classic DE algorithm to solve. The objective here is to find an optimal solution without requiring explicit knowledge of the mathematical expression or derivatives of the objective function. Classic DE methods rely on gradient information, but this is not a reliable indicator of the global optimum. Striking a balance between exploration and exploitation is challenging in non-convex functions, as large regions may have optimal solutions. Adaptability is crucial for navigating complex landscapes, as fixed parameter values can lead to premature convergence.

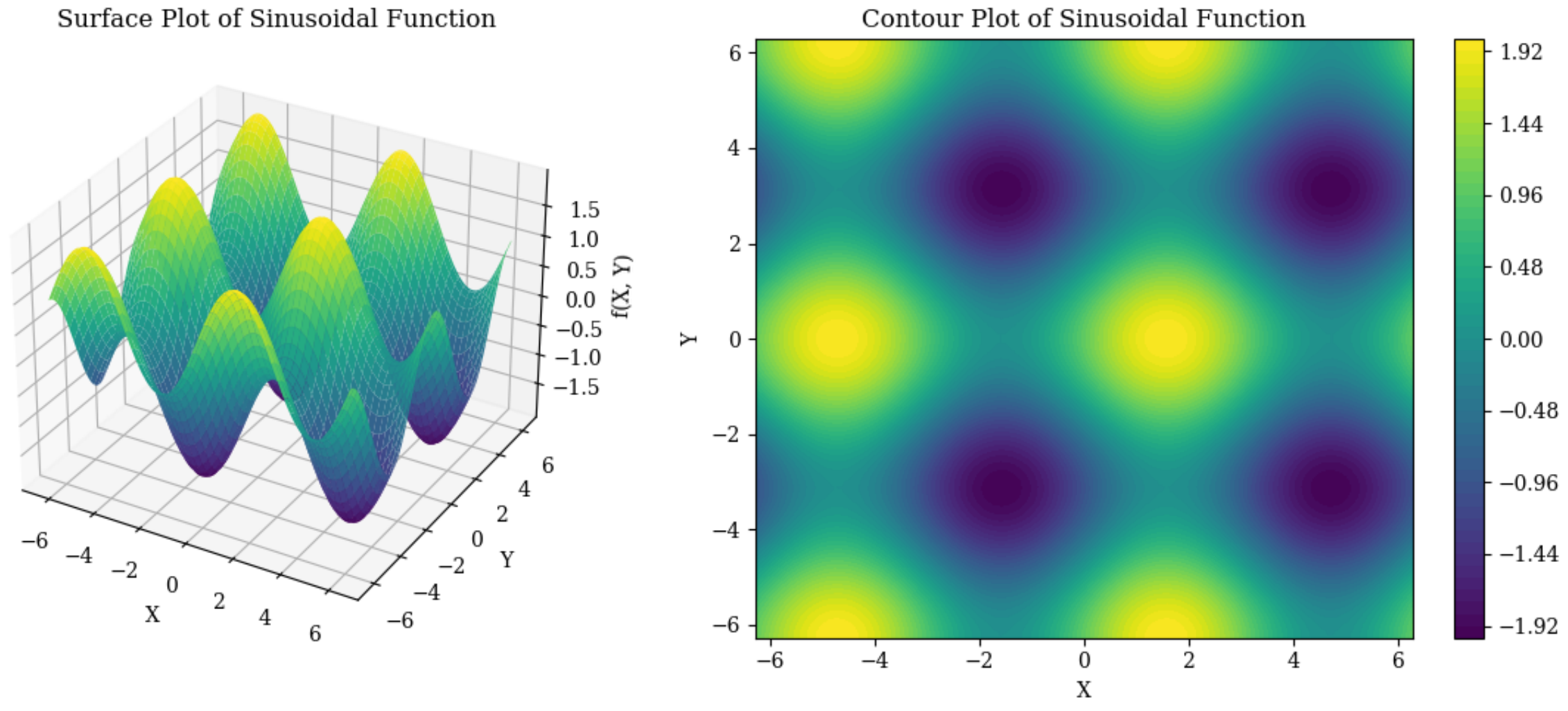

Fig. 1. Non-Convex Sinusoidal Function Landscape.

Table 1 displays the limitations of classic DE which the proposed ADED variant aims to overcome.

Table 1. Problems formulations.

| **Problems identified** | **DE issues** | **ADED approach** |
|---|---|---|
| Parameter Adaptation | The choice of control parameters (for example, mutation factor F and crossover-rate CR) can have a considerable impact on DE's performance, and obtaining acceptable parameter values is sometimes difficult. | ADED employs a basic parameter adaptation technique, modifying F and CR during the optimization phase based on the fitness and average fitness of the population. |
| Dynamic Neighborhood Topology | A fixed neighborhood topology is typically used, and all individuals interact with each other. | Introduces dynamic neighborhood topology, where individuals have varying sets of neighbors throughout the optimization process. This can enhance exploration and exploitation capabilities. |
| Limited Exploration and Stagnation | In some cases, may struggle with limited exploration of the search space and get stuck in local optima or exhibit slow convergence. | The dynamic neighborhood topology introduces variations in interactions among individuals, potentially helping to overcome stagnation and encouraging exploration. |
| Non-Convex Objective Functions | Struggle with non-convex, multimodal objective functions. | By introducing dynamic neighborhoods, ADED attempts to address challenges posed by non-convex landscapes, enhancing the algorithm's ability to navigate complex, multi-peaked fitness landscapes. |
| Robustness and Versatility | The performance can be sensitive to the control parameters used, making it less adaptable across problem domains. | It aims to provide a more robust and versatile optimization approach by adapting its parameters and introducing dynamic neighborhood topology. |

The ADED algorithm involves creating a population of individuals with "n" dimensions, assigning random values to each dimension, and adjusting the mutation rate and crossover-rate based on generation progress. It uses mutation strategies and crossover operations to calculate trial solutions and combines them with the original solution to create new candidate solutions. The algorithm stops when it reaches the maximum number of generations or fitness values stagnate for a specified number of consecutive generations. The development process of ADED is explained below.

- **Initialization**: During the initial phase of initialization, a series of basic parameters are determined according to problems. These include population number, spatial dimension, number of iterations, mutation factor (F), crossover-rate (CR), search space upper $x_{j,i}^{max}$and search space lower $x_{j,i}^{min}$. Initial population is randomly generated: $Population = [x_1, x_2, \ldots., x_{population_size}]$, each $x_1$is a vector in the search space with n dimensions. Table 2 displays the pseudocode of population initiation.

Table 2. Pseudocode for population initiation.

```
Function initialize_population(population_size, bounds):
   population = List ()

   For i = 1 to population_size:
      individual = []

      For j = 1 to length(bounds):
         // Generate a random value within the specified bounds for dimension 'j'
         random_value = RandomUniform(bounds[j]. low, bounds[j]. high)
         Add random_value to individual

      Add individuals to the population

   Return population
```

- **Fitness landscape and adaptive mutation rate**: The fitness landscape is normally used to assess the complexity of an optimization problem. Though it is instrumental in exploration and algorithm selection, the fitness landscape is not typically used as a direct basis for creating the algorithm. There are many unique features in different parts of the fitness landscape, such as steep peaks, flat plateaus, or local optima. Logically, an algorithm may be able to better explore and utilize the search space if it adjusts its mutation method in response to these landscape traits. To identify the best fit, many search strategies are developed for various fitness landscape situations; no single strategy is appropriate for all fitness landscapes. Therefore, it is crucial to devise a plan to adaptively direct population evolution depending on the fitness landscape.

  The mutation operator F is a scaling factor which controls the amplification of the differential variation between two parent individuals. Larger F enhances global search ability but slows down convergence. Smaller F accelerates convergence speed but may lead to stronger local search ability. In classic DE algorithm, F is constant throughout the optimization process. To balance global search and convergence speed, ADED starts with a larger F in the early stages of the algorithm and gradually decreasing it for faster convergence in later stages. Accordingly, ADED proposes an adaptive mutation technique based on the fitness landscape. Table 3 displays the pseudocode of adaptive mutation.

$$F = initial_mutation_rate * \left(1 - \frac{generation}{max_generation}\right)$$

According to the formula, a new adaptive mutation operator is employed where the mutation rate starts at the initial rate and declines linearly over generations, becoming closer to zero as the number of generations rises. The goal is to encourage both exploitation and exploration at different phases of the optimization process, especially exploration at the early stage and exploitation at the later stage.

Table 3. Pseudocode for adaptive mutation.

```
Function adaptive_mutation_rate (generation, max_generations,
initial_mutation_rate):
    // adaptive mutation rate based on generation and max_generations
    mutation_rate = initial_mutation_rate * (1.0 - generation /
max_generations)

    Return mutation_rate
End Function
```

The Fitness Distance Correlation (FDC) metric is used here to evaluate the correlation between the fitness values of solutions in a population and their respective distances in the search space. It provides insights into how the algorithm explores and adapts across the solution space. ADED aims to adaptively adjust parameters and explore diverse regions of the search space. FDC contributes to the adaptability of ADED in handling various optimization challenges. Fig. 2 displays the fluctuation in FDC values between high and low values across generations, which indicates dynamic behavior in the optimization process. This shows that the algorithm is adapting its strategy over generations. It also shows a cyclic pattern of convergence and divergence.

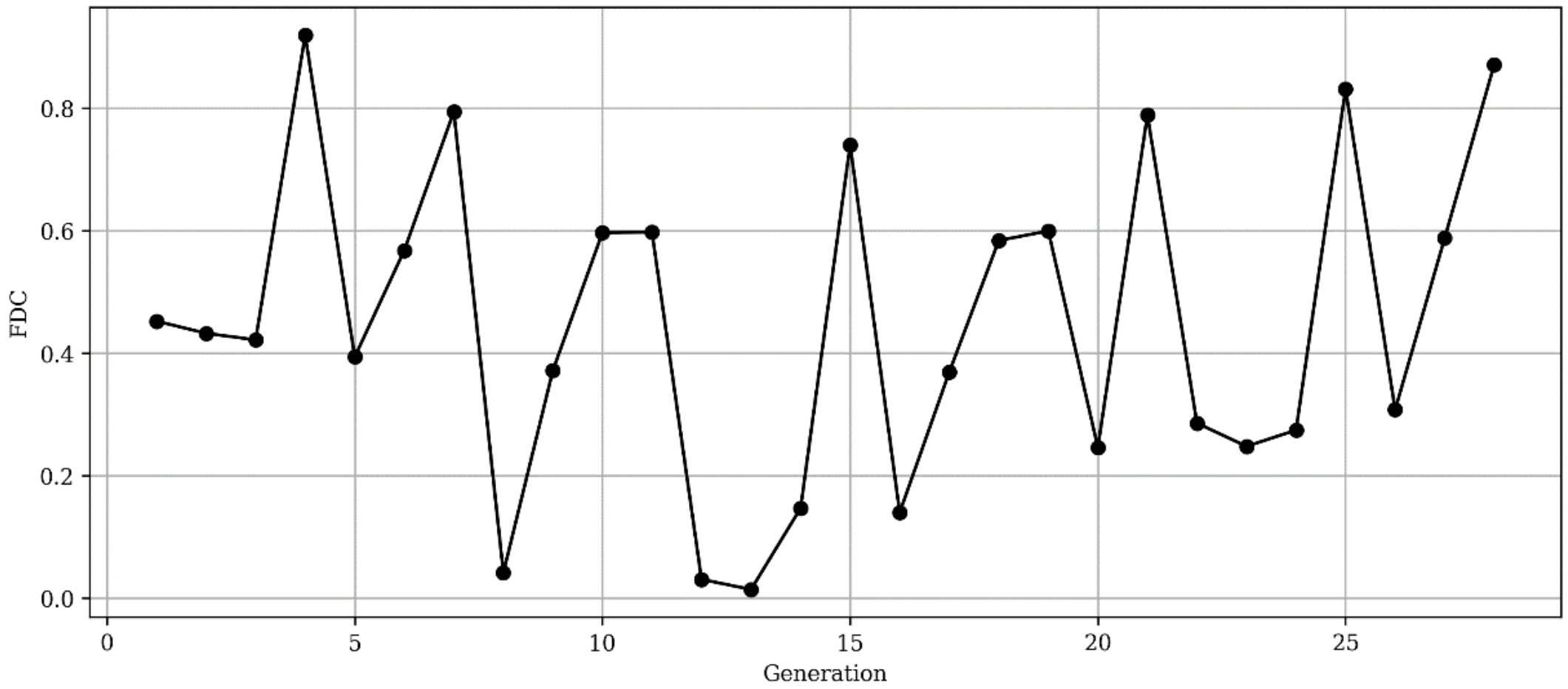

Fig. 2. Fitness-Distance Correlation (FDC) across generations.

The main elements of the ADED algorithm's dynamic neighborhood updates are provided in Table 4. By eliminating itself and modifying the neighborhood size according to the total number of individuals and the existing neighborhood size, this function makes it easier for a given individual to be randomly selected as a neighbor. By updating a specific neighborhood in response to a trial solution's fitness, the update_neighborhoods function improves the algorithm's exploration-exploitation approach. With this function, the algorithm would be able to maneuver intricate solution spaces and adjust to various environments.

The set of all individuals excluding the current individual: $all_individuals = \{0, 1, 2, \ldots, i-1, i+1, \ldots, N-1\}$. The neighborhood size: $neighborhood_size = min(k, N-1)$

- N be the total number of individuals in the population.

- i be the index of the individual for which the neighborhood is being updated.
- k be the existing neighborhood size for the individual at index i.

$$dynamic_neighborhood(i, neighborhoods) = random_choice(all_individuals, neighborhood_size)$$

$$update_neighborhoods(i, neighbors, f_{trial}, neighborhoods) = for\ neighbor_index\ in\ neighbors: if\ f_{trial} < f_{neighbor} : neighborhoods[i][neighbor_index] = f_{trial}$$

- $f_{trial}$ = fitness of trial solution
- $f_{neighbor}$ = fitness of a neighbor
- For each neighbor in the selected set of neighbors, if $f_{trial} < f_{neighbor}$, update the neighborhood with the trial solution's fitness.

Table 4. Pseudocode of dynamic neighborhood update in ADED.

```
function dynamic_neighborhood(individual_index, neighborhoods):
    all_individuals = range(len(neighborhoods))
    all_individuals.remove(individual_index)
    neighborhood_size = min(len(all_individuals), len(neighborhoods[individual_index]))
    neighbors = random_choice(all_individuals, neighborhood_size, replace=False)
    return neighbors

function update_neighborhoods(individual_index, neighbors, trial_fitness, neighborhoods):
    for neighbor_index in neighbors:
        neighbor_fitness = evaluate_fitness(neighborhoods[individual_index][neighbor_index])
        if trial_fitness < neighbor_fitness:
            neighborhoods[individual_index][neighbor_index] = trial_fitness
```

- **Adaptive crossover-rate** $(CR) = initial_crossover_rate * \frac{generation}{max_generation}$. The pseudocode for adaptive crossover is presented in Table 5.

Table 5. Pseudocode for adaptive crossover-rate.

```
Function adaptive_crossover_rate(generation, max_generations, initial_crossover_rate):
    // adaptive CR based on generation and max_generations
    crossover_rate = initial_crossover_rate * (generation / max_generations)

    Return crossover_rate
End Function
```

- **Mutation method:** The algorithm begins by using the usual $DE/rand/1$ mutation strategy, which entails scalarizing the differences between two randomly selected individuals from the population to create a trial solution. Here, the mutation strategies are used to compute the trial solution $v_i$ for each candidate solution $x_i$. By producing fresh candidate solutions based on the differences between randomly chosen population members, this approach aids in the exploration of the solution space. The process of mutation is made more varied by the randomization in neighbor selection, which enables the algorithm to explore a wider range of search spaces.

$$v_i = x_{r1} + F * (x_{r2} - x_{r3})$$

$r1, r2, and\ r3$ are the distinct random indices representing different solutions in the population, and F is the adaptive mutation factor.

- **Crossover operation**: The classical Differential Evolution (DE) algorithm traditionally employs two constant crossover-rate (CR) methods (Zhu et al., 2020). However, the limitation lies in the constant value of CR, neglecting considerations for both global search ability and search speed.

  In the binomial crossover, a crucial aspect of the evolution of target individuals $x_i$ involves ensuring that the components of at least one dimension of trial individuals $u_i$ are generated by the mutation individual $v_i$. Meanwhile, the components of other dimensions are determined by the CR parameter in a random manner. The proposed ADED algorithm addresses this limitation by combining the trial solution $v_i$ with the original solution $x_i$ to generate a new candidate solution $u_i$. In ADED, the crossover rate is not a fixed constant but adapts dynamically, contributing to both global search capability and search speed. This dynamic adaptation of CR, along with other enhancements in ADED, aims to improve the exploration-exploitation trade-off and the algorithm's overall performance in optimization scenarios.

Table 4 displays pseudocode combining mutation strategy and crossover operation.

$$u_i[j] = \begin{cases} v_i[j] & if\ \ j = rand(0,1) < CR\ or\ j = randint(1, D) \\ x_i[j] & otherwise \end{cases}$$

- $u_i[j]$ is the j^th^ component of the trial vector for the i^th^ individual.
- $v_i[j]$j] is the j^th^ component of the mutant vector for the i^th^ individual.
- $x_i[j]$] is the j^th^ component of the target vector for the i^th^ individual.
- $rand(0,1)$ is a random number between 0 and 1.
- $randint(1, D)$ generates a random integer between 1 and D (the dimensionality of the vectors).
- CR is the crossover rate, determining the probability of crossover occurring for each component.

The above expression indicates that for each component j, there are two possible outcomes based on the comparison with a random number and the crossover rate. $if\ \ j = rand(0,1) < CRor\ j = randint(1, D)$ then the $j^{th}$ component of the trial vector $u_i$ is assigned the value of the $j^{th}$ component of the mutant vector $v_i$. Otherwise, the $j^{th}$ component of trial vector $u_i$ is assigned the value of the $j^{th}$ component of the target vector $x_i$. This method incorporates information from the mutant vector into the trial vector in a probabilistic and selective manner, based on both a random decision and the predetermined crossover rate. The term "binomial" alludes to the binary nature of each component's decision.

1) The expression for exponential crossover is represented using an exponential decay function. Here, exponential crossover occurs, where the trial vector is formed by combining the information from the target and mutant vectors in an exponential manner. This method incorporates information from the mutant vector into the trial vector in a probabilistic and selective manner, based on both a random decision and the predetermined crossover rate.

Table 6 provides the pseudocode for mutation strategy and crossover operation.

Table 6. Pseudocode Mutation strategy & crossover operation:
F is dynamically adjusted based on the generation progress using the adaptive mutation rate function.

```
For i from 0 to population_size - 1:
    individual = population[i]

    // Adaptive mutation rate and crossover rate
    F = adaptive_mutation_rate(generation, max_generations)
    CR = adaptive_crossover_rate(generation, max_generations)

    neighbors = list(range(population_size))

    // Randomly select two distinct neighbors
    neighbor1 = population[select_random_element(neighbors)]
    neighbor2 = population[select_random_element(neighbors, exclude=neighbor1)]

    // Trial solution using DE mutation strategy
    trial_solution = individual + F * (neighbor1 - individual) + F * (neighbor2 - individual)

    // Fitness of the trial solution
    trial_fitness = objective_function(trial_solution)

    // Update the population with the trial solution if it's better
    if trial_fitness < fitness[i]:
        population[i] = trial_solution
        fitness[i] = trial_fitness

End For
```

Finding the global optimum in a complex fitness landscape is a challenging task. ADED aims to approximate the optimal solution by iteratively updating the population based on the defined strategies. The concept of global fitness is associated with the best fitness value encountered during the optimization process. The algorithm evaluates the fitness of candidate solutions using the objective function, with the goal of minimizing or maximizing this function, and the best fitness value achieved by the algorithm is considered an approximation to the global optimum.

- **Local Search**: After generating a trial solution, a local optimization algorithm is applied to refine and potentially replace it with a better solution. ADED aims for a balance between exploration and exploitation, incorporating local search. This choice aligns with the overall goal of finding near-optimal solutions in the optimization process. ADED implements L-BFGS-B (Limited-memory Broyden-Fletcher-Goldfarb–Shanno with Bounds), which is a quasi-Newton optimization method that handles bound constraints. Table 5 displays the pseudocode for local search.

$$x_i = local_optimization(v_i)$$

Table 7. Pseudocode for local search.

```
Function local_search(solution):
    result = minimize (objective_function, solution, method='L-BFGS-B', bounds=bounds)

    // Return the optimized solution
    Return result.x
End Function
```

- The candidate solutions for the next generation are selected based on their fitness values and with lower fitness values.

$$Population = select_new_population(population)$$

- During the monitoring of the convergence of the optimization process, the algorithm stops when either of the following conditions are met:
    - It reaches the maximum number of generations specified by max_generations.
    - Fitness values have stagnated for stagnation_limit consecutive generations.
    - These criteria help prevent the algorithm from running indefinitely and provide a mechanism for early stopping if convergence is achieved.

Table 8 displays the pseudocode of early stopping criteria if convergence is achieved.

Table 8. Pseudocode for convergence threshold.

```
Function has_converged(best_fitness_history, stagnation_limit):
    // Check if the length of best_fitness_history is less than the stagnation_limit
    If length(best_fitness_history) < stagnation_limit Then
        Return False // Convergence not reached.
    End If

    // Check if the last 'stagnation_limit' elements in best_fitness_history are all the same
    If all(best_fitness_history[-stagnation_limit:] == best_fitness_history[-1]) Then
        Return True // Convergence has been reached.
    Else
        Return False // Convergence not reached.
    End If
End Function
```

- The optimal solution to the problem is generally regarded as being the best solution discovered during the optimization process.

Directly expressing the distribution as a simple analytical formula is often infeasible due to the complexity of real-world optimization landscapes. Instead, the invariant distribution of solutions is determined empirically through the optimization process itself. The algorithm explores the solution space by iteratively generating and evaluating candidate solutions, adapting its parameters, and gradually converging towards regions of interest in the search space.

The sinusoidal function is chosen here because it is non-convex and has multiple local optima. Sinusoidal functions exhibit complex and oscillatory behavior, providing a diverse landscape of peaks and valleys. This complexity makes the optimization problem more interesting and tests the algorithm's ability to explore and exploit the search space. Moreover, sinusoidal functions have a known global optimum and multiple local optima. This helps to assess whether ADED can efficiently locate the global optimum while avoiding the local optima trap.

Fig. 3 provides a visual display of where the algorithm found solutions in the search space defined by the x and y axes. Areas with a higher frequency of solutions (elevated regions in the 3D plot) indicate that the algorithm frequently finds solutions in those areas. This suggests that those regions contain multiple local optima. The peak that stands out from the rest suggests the location of the global optimum. Multiple peaks in the plot indicate the presence of local optima, and the algorithm has found solutions in these local optima.

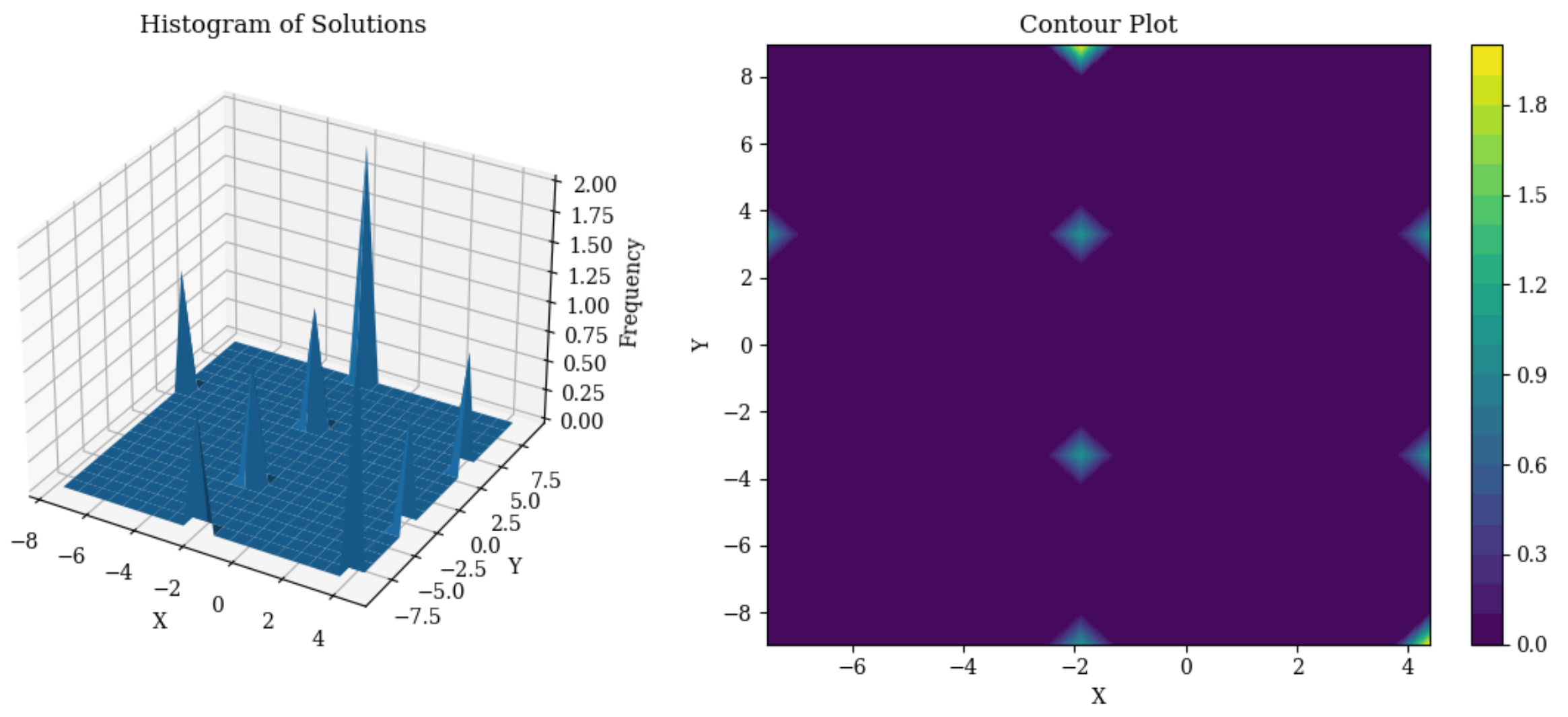

Fig. 3. Solution Distribution and Density.

The ADED algorithm is configured here to run with a population size of 50 and a maximum of 100 generations, and it is restricting the search space to the range (bounds) of [-10, 10]. The algorithm tries to find the optimal solution within these constraints.

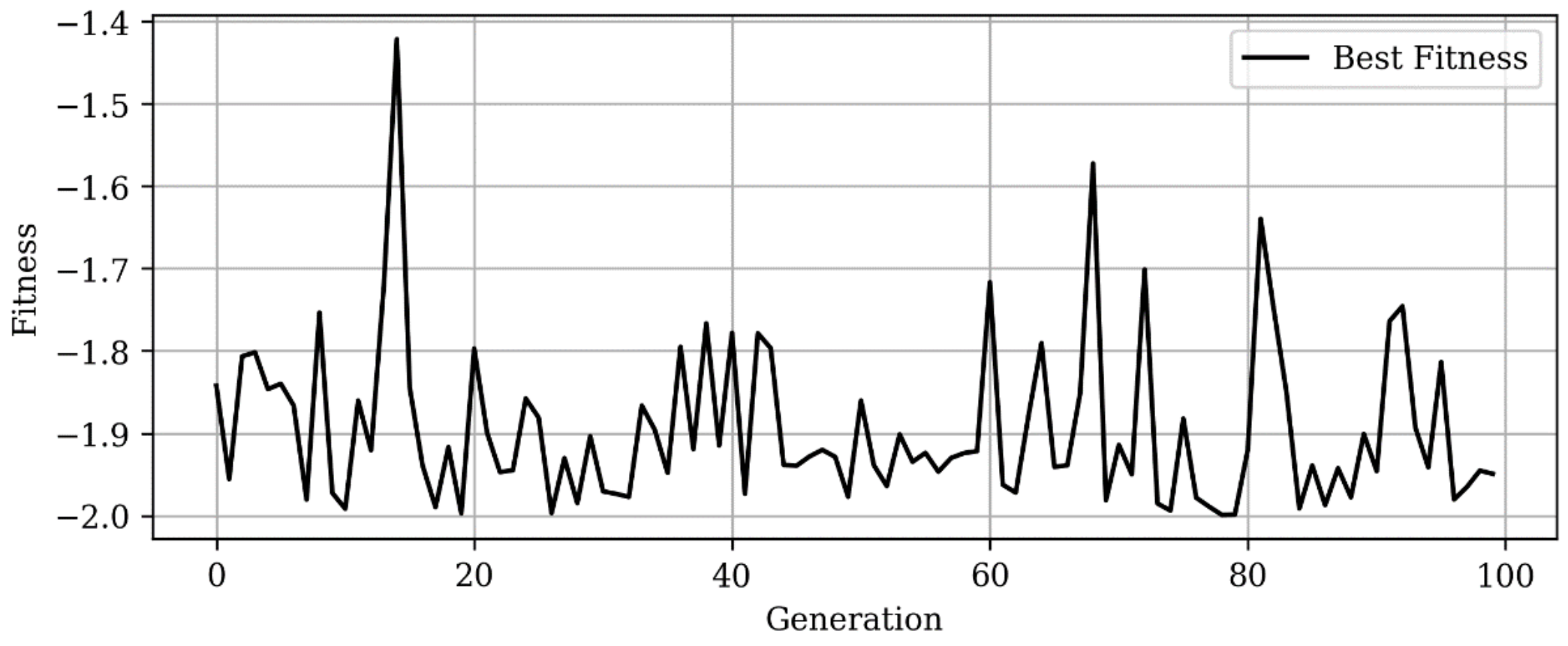

Fig. 4. ADED Convergence plot: Sinusoidal objective function.

Fig. 4 displays the convergence, where we see that the fitness value decreases as the number of generations increases. This indicates that the optimization algorithm is converging toward a better solution over time. The decreasing trend suggests that the algorithm is effective in finding improved solutions at each generation.

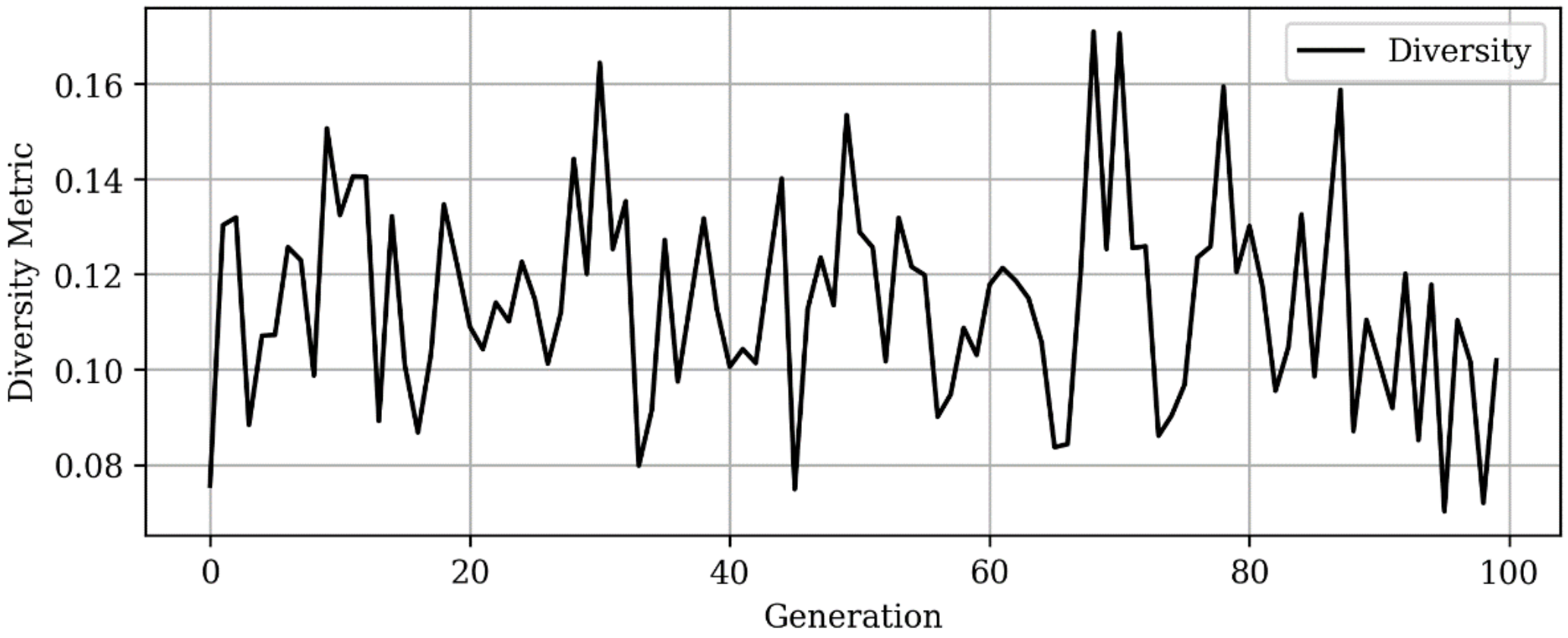

Fig. 5. Diversity: ADED with Sinusoidal objective function.

Monitoring diversity is crucial. Fig. 5 displays diversity and how the diversity metric evolves over generations in optimization. The initial increase in diversity, followed by oscillations, indicates a balance between exploration and exploitation. The initial increase from 0.08 to almost 0.13 suggests that the algorithm introduced diversity in the population during the early generations. The declining trend at the end suggests that the algorithm is converging, which is expected as the optimization process aims to refine solutions towards optimal ones.

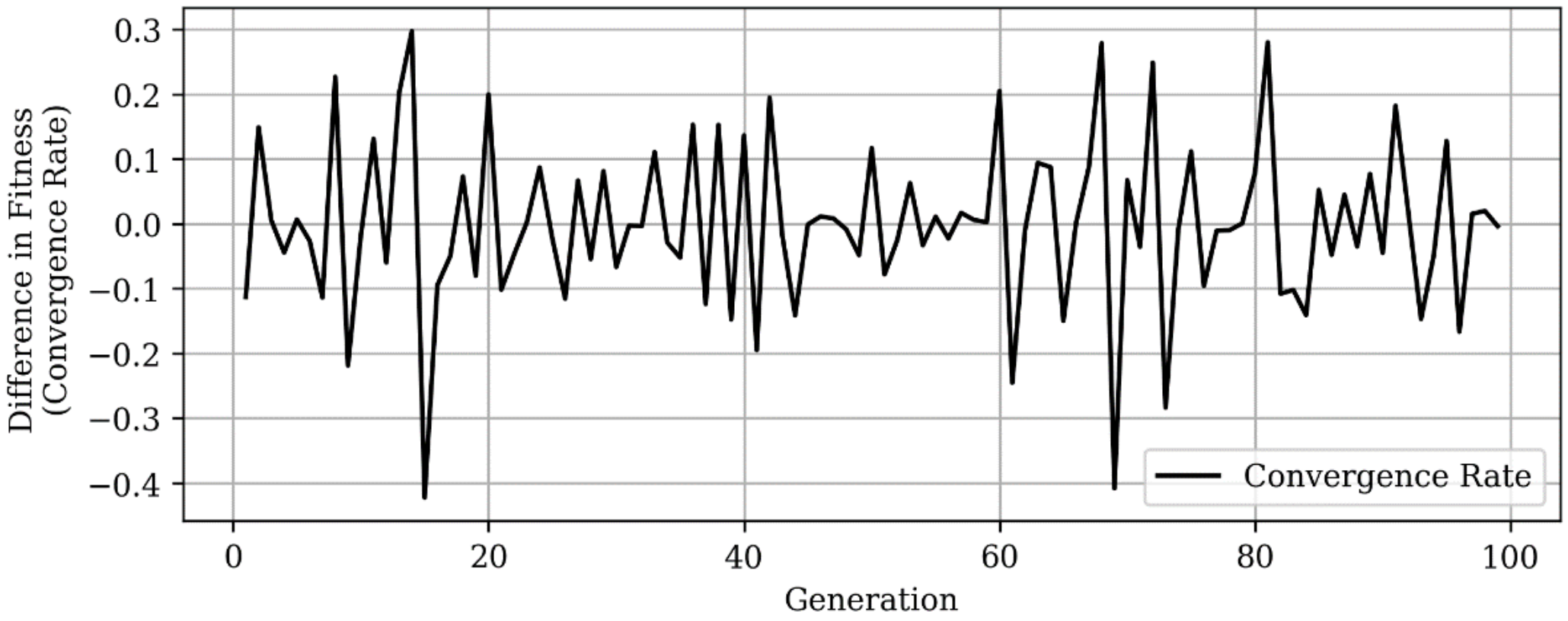

Fig. 6. Convergence Rate.

Analyzing the convergence rate helps understand how quickly the algorithm is improving. Fig. 6 displays the convergence rate, showing the improvement of the algorithm, with negative values indicating decreasing fitness values and larger values indicating faster convergence. As the process progresses, significant improvements become less frequent, indicating the algorithm is reaching the local optimum.

These three plots collectively illustrate the optimization process's dynamics and efficiency. The convergence plot demonstrates that the algorithm is consistently finding better solutions over time. The time taken for the whole run is just 95.61 seconds, which shows a fast execution time. The convergence rate plot quantifies the speed of improvement, with rapid convergence at the early stages of optimization.

Table 9 displays the optimal solution and convergence rate of ADED, which is tried with both: (1) dynamic neighborhood topology, where the set of neighbors for each individual changes over time. Instead of having a fixed set of neighbors for each individual throughout the optimization process, a random subset of individuals is chosen as neighbors for each interaction. This randomness introduces variation in the neighborhood structure, allowing individuals to explore different regions of the solution space by interacting with different sets of neighbors. (2) No Dynamic Neighborhood Topology (All Individuals Are Neighbors). Here, every individual is considered a neighbor of every other individual throughout the entire optimization process. The neighborhood structure remains constant, and each individual interacts with all other individuals in the population. Because every individual has the capacity to influence or be impacted by any other individual, this setup led to a more global investigation of the solution space. The choice of neighborhood topology can influence the exploration-exploitation trade-off in the optimization process.

Table 9. Optimal solution and convergence rate.

| **Neighborhood Topology** | **Parameters** | **Best solution** | **Best fitness** | **Final convergence rate** | **Execution time (seconds)** |
|---|---|---|---|---|---|
| No Dynamic Neighborhood Topology | population size = 50<br>max generations = 100<br>bounds = [(-10, 10)] | [ -7.8588, 40.8507, -90.8911, 33.1912, 308.4996, 150.4668, 14.6875, 56.7643, 478.3643, -16.0947] | -1.9999 | -0.17116 | 95.61 |
| Dynamic Neighborhood Topology | | [-26.6864, 15.7097, -1.9141, -12.8803, 6.7110, 15.5982, 10.8560, 0.1927, 7.9588, -12.7255] | -2.0000 | -0.26945 | 76.88 |

The algorithm finds a low fitness value (close to -2.0) for the best solution, indicating a near-optimal or optimal solution. The negative final convergence rate (-0.17116) suggests a decrease in fitness values towards the end of the optimization process, indicating convergence to a solution. The chosen parameters, such as a population size of 50 and a maximum of 100 generations, were used for the optimization run. The execution time of 95.61 seconds provides an indication of the computational resources required for the optimization process. The final convergence rate was close to zero, indicating a stabilized search. The best fitness value is close to the global optimum, indicating successful performance. Overall, it suggests that ADED successfully identified a solution with a fitness close to the optimal value within the specified parameter and computational constraints. Fig. 7 displays the convergence behavior of Classic DE and ADED with 10 runs when applied to a convex objective function.

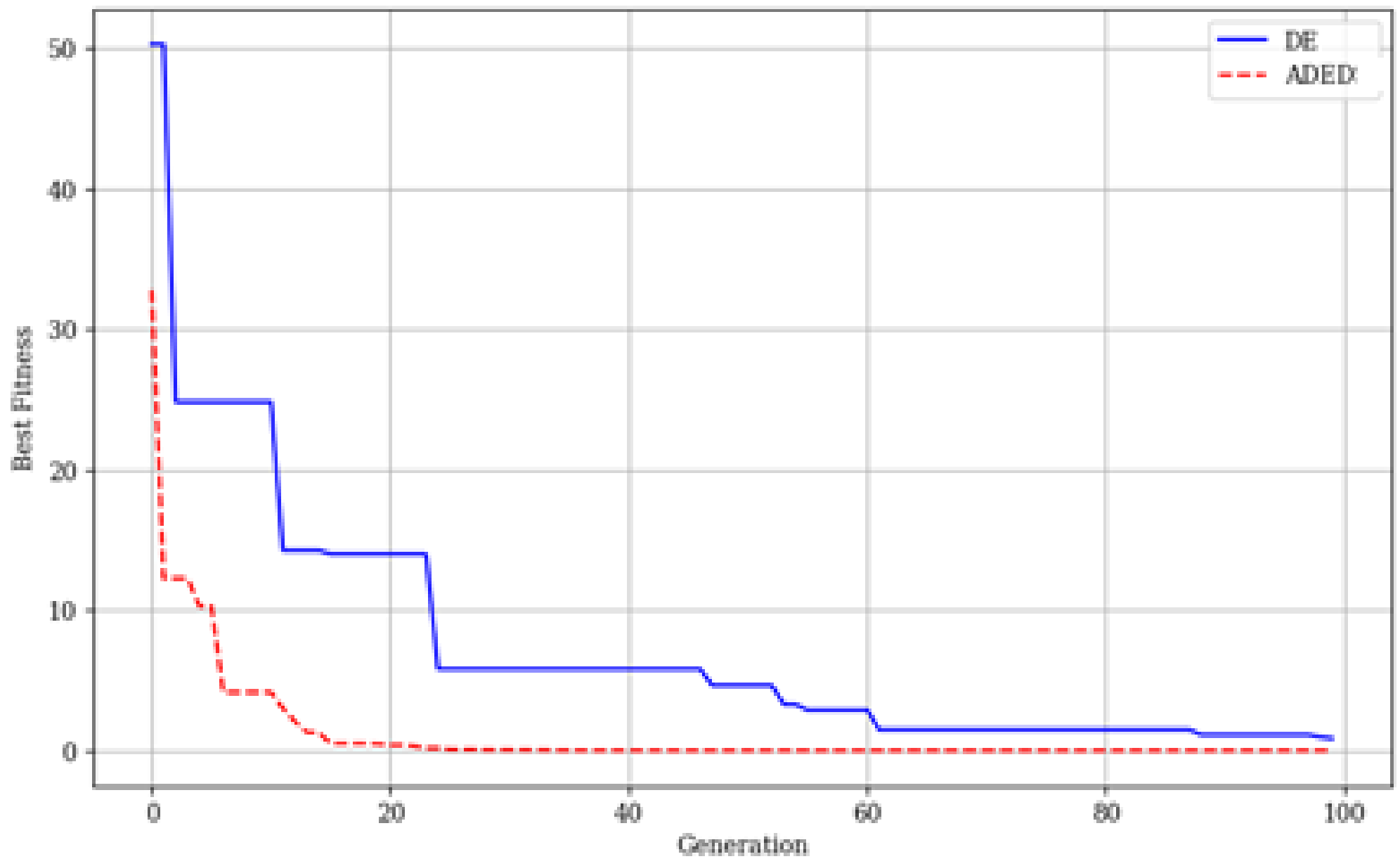

Fig. 7. Convergence plot - Classic DE & ADED.

The algorithm that converges to a lower fitness value faster is considered more efficient. We see from Fig. 7 that the ADED line is lower than the DE line, indicating the ADED algorithm found a better solution with the same run and configuration.

The plots for different runs of ADED (Fig. 8) show similar patterns, suggesting that the algorithm's performance is relatively consistent. Multiple runs help assess the ADED's behavior across different scenarios.

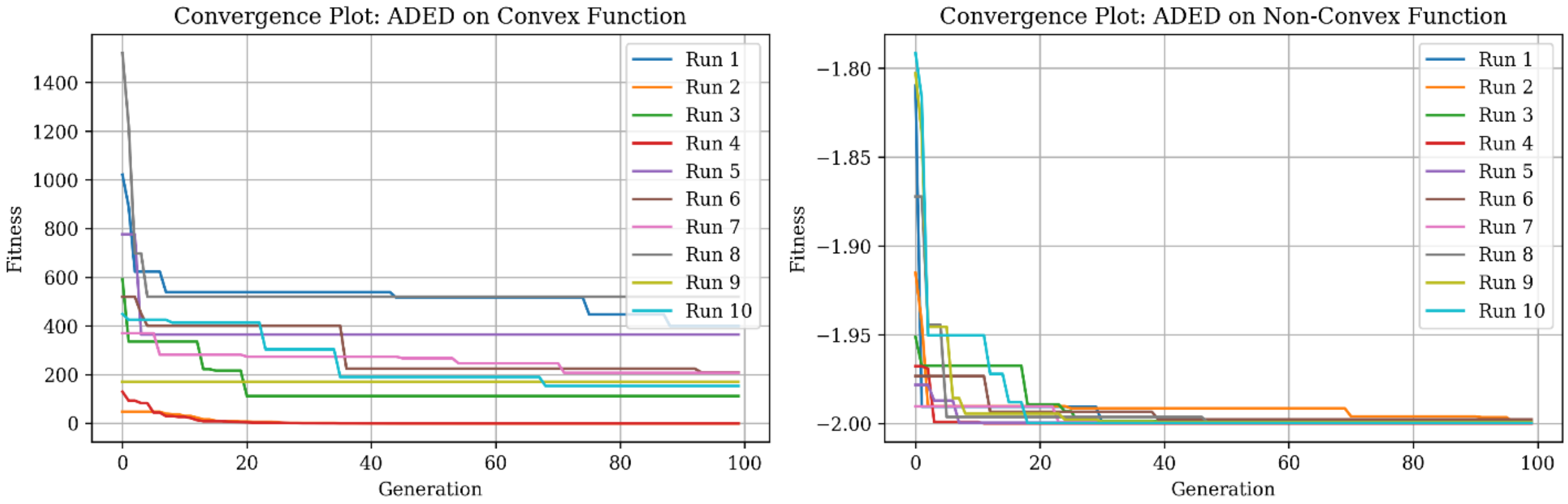

Fig. 8. ADED Convergence plots of multiple runs.

There is some variability in the convergence plots across runs, indicating sensitivity to initial conditions or stochastic elements. Stochastic elements, such as random selection of individuals, contribute to variability between runs. This variability is expected in non-convex sinusoidal optimization problems and highlights the challenge of navigating complex, multi-modal landscapes.

Fig. 9 displays convergence patterns plotted for the non-convex function with different mutation rates (F values).

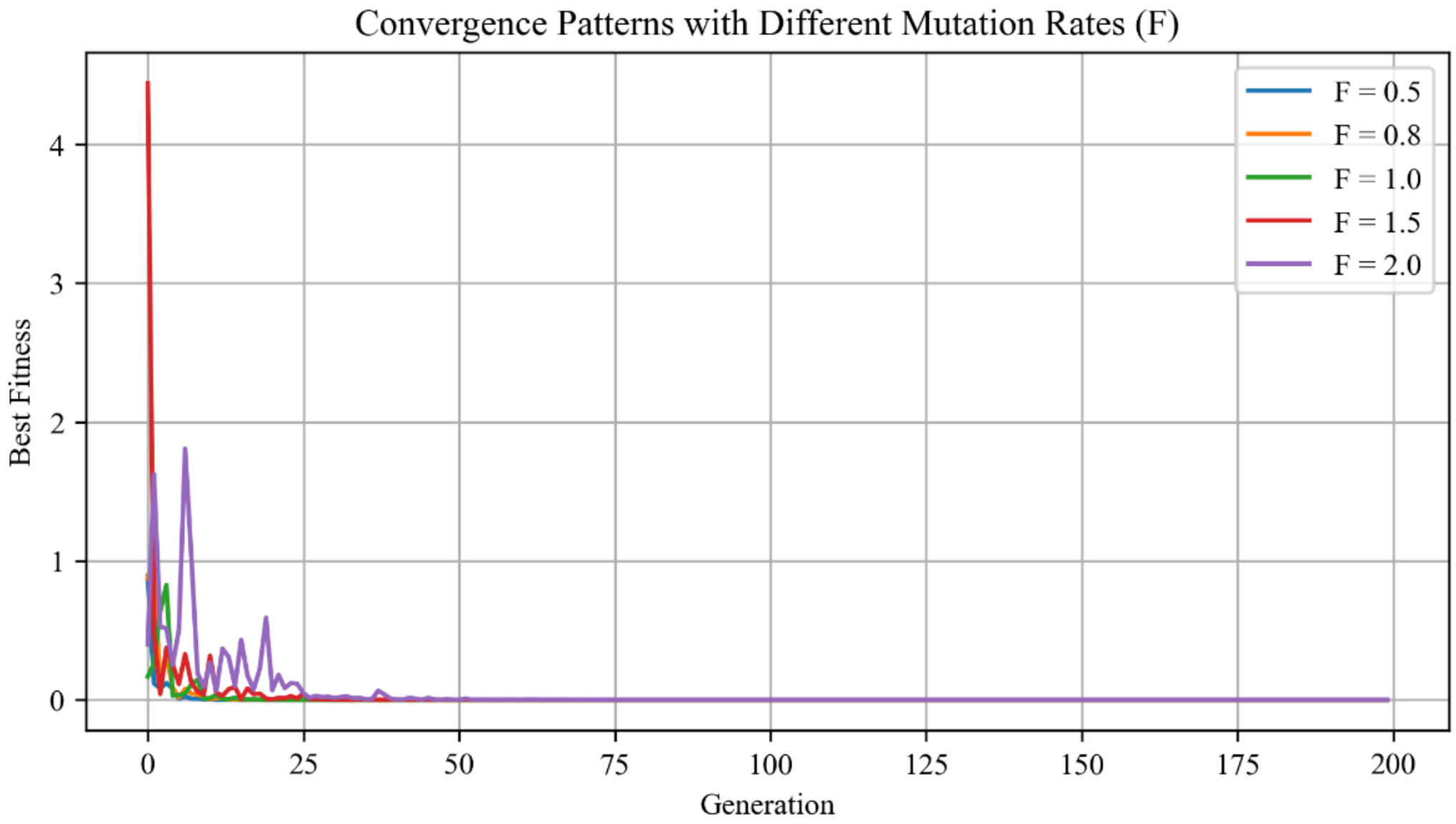

Fig. 9. ADED Convergence pattern for convex function with different F values.

Smaller F values (e.g., F = 0.5, 0.8) result in slower convergence and a more cautious exploration of the search space. Larger F values (e.g., F = 1.5, 2.0) lead to faster exploration and potentially jumping over local optima, but they also introduce more noise or oscillations in the convergence pattern. As can be seen from Fig. 9, when the mutation factor (F) is set to a higher value (e.g., F = 1.0 to 2.0), it allows for a more significant influence of the mutant vector in generating trial vectors. This increased influence led to larger steps in the search space during mutation, encouraging exploration of new regions. However, larger steps result in overshooting the optimal solutions, leading to oscillations. In contrast, lower values of F (e.g., F = 0.8) result in smaller steps during mutation, which lead to more stable convergence; however, this may struggle to explore distant regions. The convergence patterns with oscillations indicate challenges in finding a stable solution. This is due to the balance between exploration and exploitation or other factors like population size. Smoother convergence patterns (less oscillation) suggest a more stable convergence to a global optimum. However, ADED's aim is to find the global minimum efficiently in more complex and non-convex landscapes. Thus, to better understand the shortcomings of Classic DE, we compare the behavior elaboratively with Fig. 10, which displays the convergence of the sinusoidal objective function.

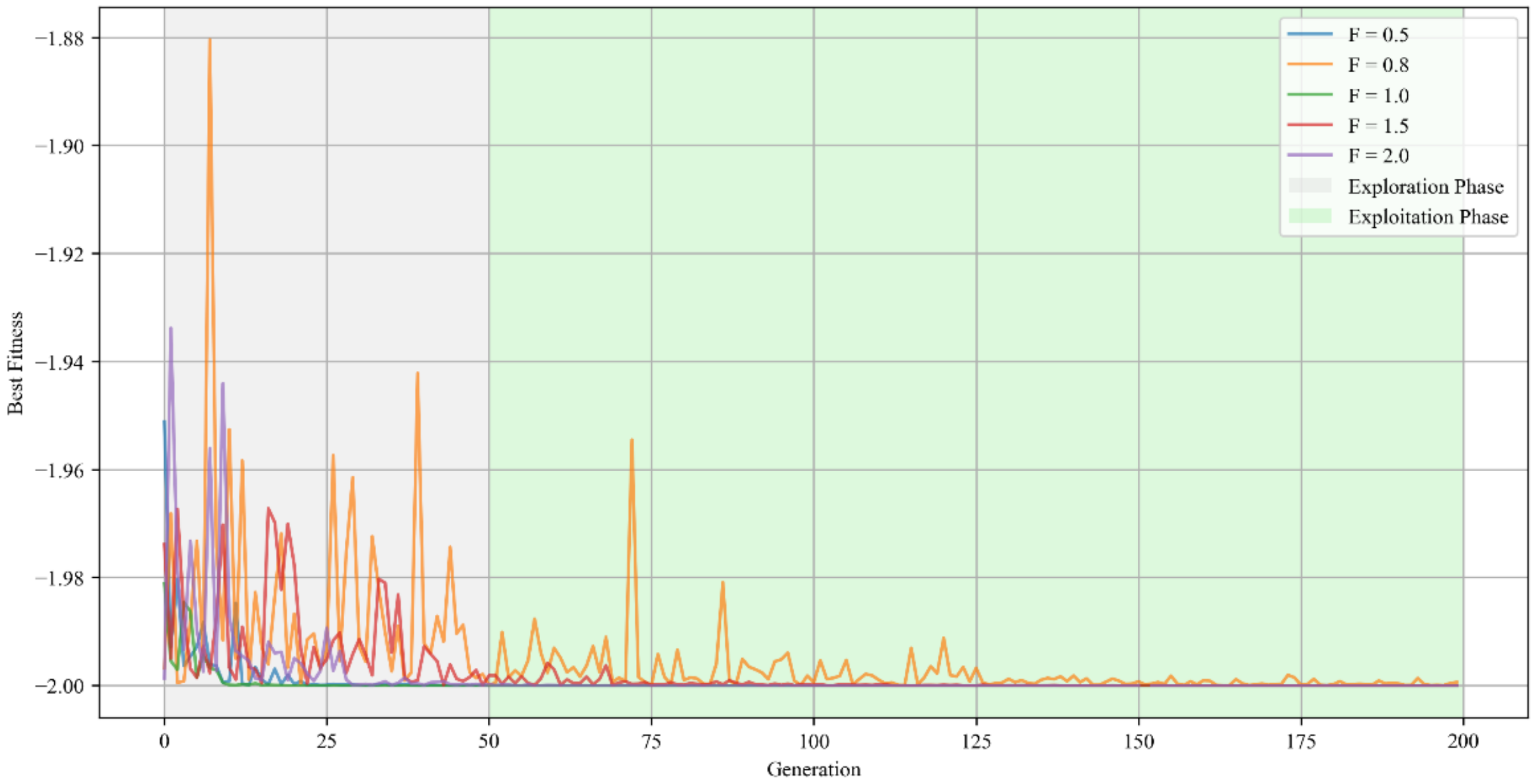

Fig. 10. Classic DE Convergence pattern for non-convex function with different mutation (F) rates.

The shaded regions in the plot indicate exploration and exploitation phases. During the early generations, especially between 0 and 50 along the x-axis (the grey-shaded exploration phase), the algorithm is exploring the search space broadly. In later generations (the green-shaded exploitation phase), it focuses on refining solutions in promising regions. Different F values can influence the duration and effectiveness of these phases. Smaller F values (0.5) prolong the exploration phase, while larger F values accelerate exploration and the transition to exploitation. Here, we see that the algorithm is struggling to find an optimal solution with a non-convex function, reflecting the inherent limitations of Classic DE.

To get a clear picture of the oscillation and challenges faced by DE in finding the best solution, we enlarged the plot using a single mutation factor (F = 0.8). Fig. 11 provides a good illustration of how DE struggles to find the best search space when applied to a non-convex optimization problem.

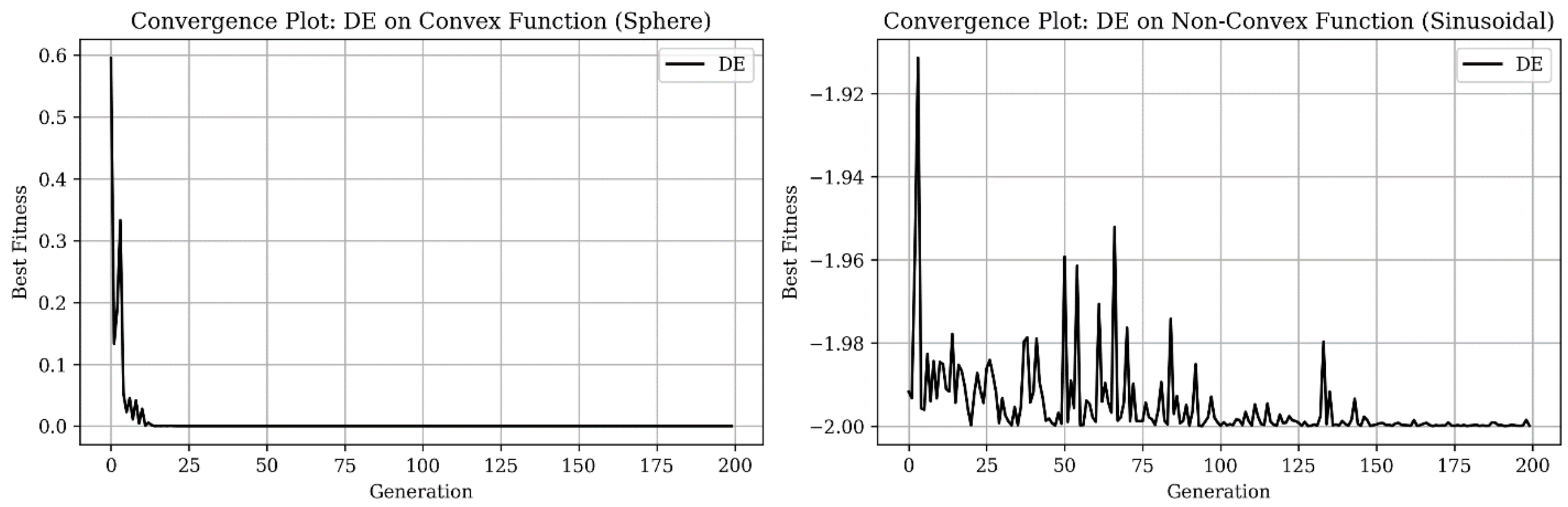

Fig. 11. Convergence plot Classic DE (F=0.8).

Fig. 12 displays the convergence behavior of ADED for both convex and nonconvex optimization problem.

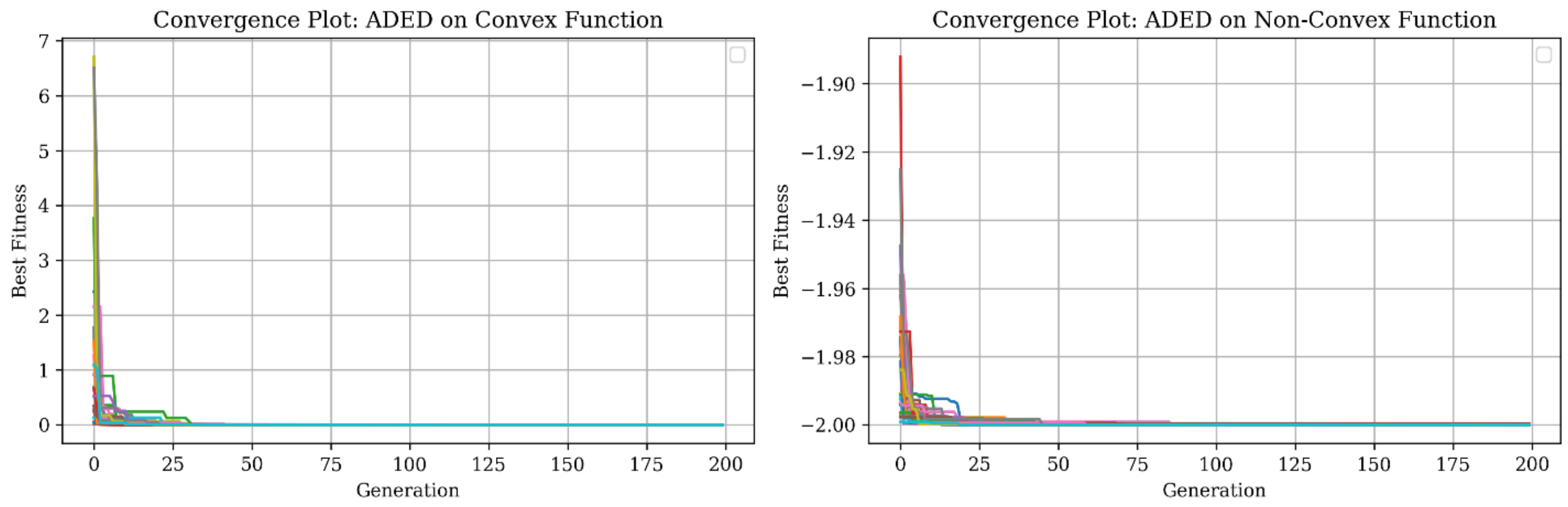

Fig. 12. Convergence behavior with ADED.

With the application of ADED, the convergence plot for the convex function shows a decreasing trend over generations, indicating that the algorithm is improving its fitness and approaching the optimal solution. The convergence plot for the non-convex function shows more variability, with multiple decreases and plateaus. This reflects the algorithm exploring different regions and converging to local minima. It takes longer to converge compared to the convex function, and the final fitness values may vary across runs.

In the case of DE, the default values of F = 0.8 and CR = 0.9 are used because these values have been found to work well for a variety of optimization problems, striking a balance between exploration (mutation) and exploitation (crossover). For ADED, the mutation rate (F) and crossover-rate (CR) are set to (0.5, 2.0) and (0.1, 0.9), respectively. This gives the algorithm versatility, allowing it to modify these rates dynamically dependent on the performance of others in the population. Empirical research (Das et al., 2015) shown that a F value between 0.5 and 2.0 enhances performance significantly, particularly for large-scale issues. It should be emphasized that while higher F values aid in exploring previously unexplored areas, relinquishing control over leap length and donor location may not be warranted. Smaller F values, on the other hand, result in an exploitative search over already explored regions, boosting exploration. Table 10 shows a comparative performance evaluation. For each variant, the table includes the algorithm parameters, the best solution found, and the corresponding best fitness value. The results are based on 30 independent runs for each optimization problem.

Table 10. Comparative Performance of Classic DE and ADED.

| Variants | F | CR | params | Best solution | Best fitness |
|---|---|---|---|---|---|
| Classic DE (convex) | 0.8 | 0.9 | population_size = 300<br>max_generations = 200<br>bounds = [(-10, 10)]<br>num_runs = 30 | [6.3954e-19, 1.5245e-19] | 4.3226e-37 |
| Classic DE (non-convex) | | | | [-158.6497, 311.0194] | -1.6879 |
| ADED (convex) | (0.5, 2.0) | (0.1, 0.9) | | [-5.2984e-10 9.3322e-09] | 2.0129e-18 |
| ADED (non-convex) | | | | [-1.5755 3.1586] | -1.9999 |

Both algorithms seem to find very good solutions with extremely low fitness values, indicating they are effective for convex optimization. For non-convex, DE obtained a fitness value of -1.6879. However, ADEDS has its adaptability and dynamic neighborhood approach, making it more robust in handling different types of optimization landscapes and finding near-optimal fitness (-1.99). The fitness values alone do not necessarily indicate a better algorithm; the speed of convergence is crucial. Running the algorithms on multiple instances of the problem provides a reasonable assessment of their robustness and consistency in finding good solutions.

The process workflow for the ADED method is depicted in Fig. 13, which includes initialization, parameter adaptation, dynamic neighborhood selection, trial solution creation, fitness evaluation, crowding selection, and neighborhood updating. The algorithm iteratively evolves a population of solutions to optimize the given objective function.

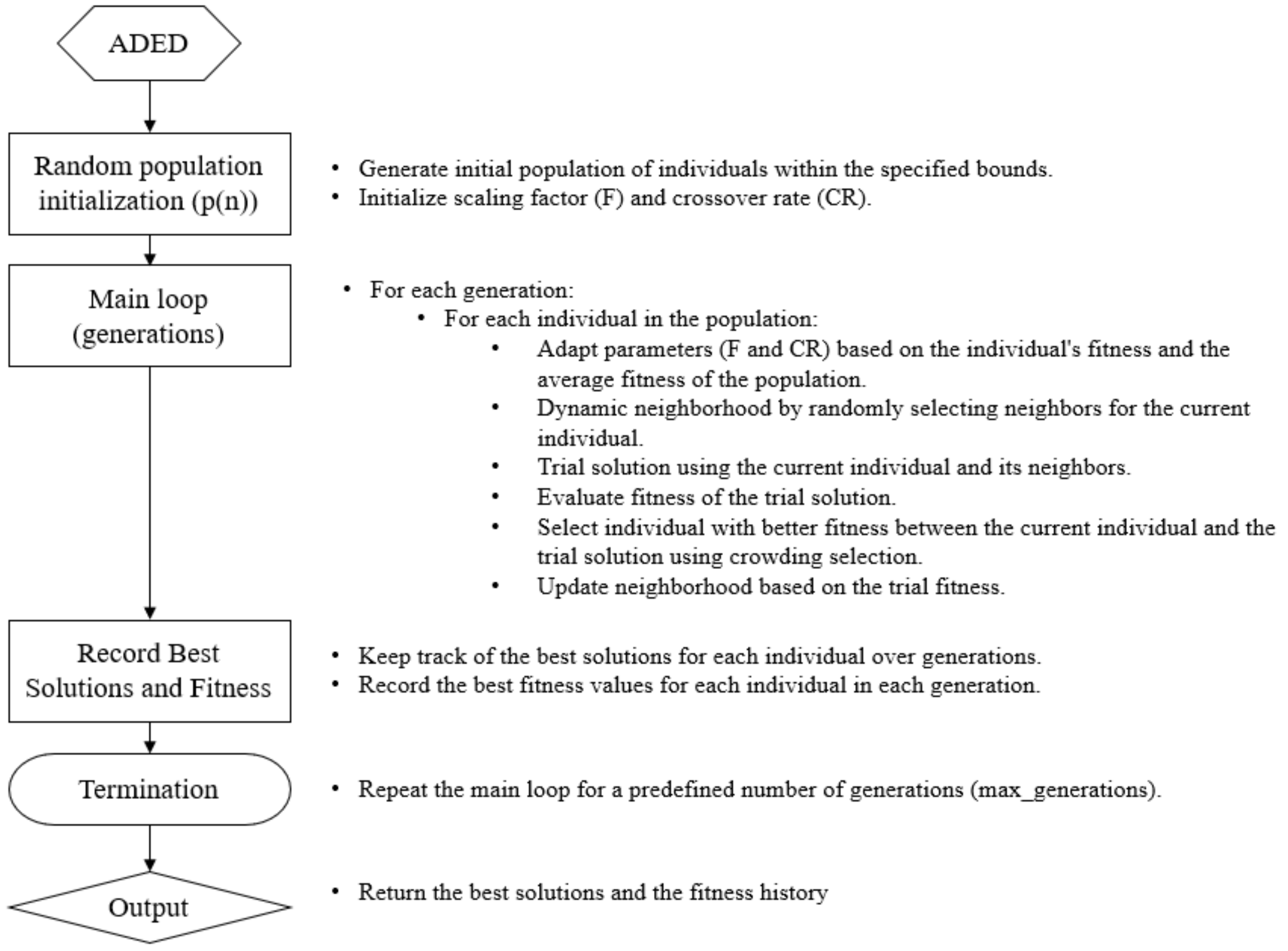

Fig. 13. ADED development workflow (Source: Author).

Table 11 displays the pseudocode which is applied to both convex and sinusoidal non-convex optimizations.

In the convex problem, the objective function is $f_{convex}(x) = x_0^2 + x_1^2$. The objective function is smooth and has a single optimal solution at the center, which DE successfully finds. In the non-convex problem, the Sinusoidal objective function is $f_{nonconvex}(x) = \sin(x_0) + \sin(x_1)$ where DE struggles to find the global minima due to the presence of multiple local optima. The non-convex function presents a landscape where the algorithm is searching for the minimum value. However, there are many points in the search space where the gradient (derivative) of the function is zero, indicating a potential local minimum.

$$\nabla f_{nonconvex}(x_0^* + x_1^*) = \left(\frac{\partial f_{nonconvex}}{\partial x_0}(x_0^* + x_1^*),\ \frac{\partial f_{nonconvex}}{\partial x_1}(x_0^* + x)\right) = (0,0)$$

while DE converged to a solution $(x_0^* + x_1^*)$, it may not be the best possible solution due to the presence of potentially better global optimum elsewhere in the landscape.

$$f(x_0^* + x_1^*) \leq \min_{global} f(x_0 + x_1)$$

Where $\min_{global} f(x_0 + x_1)$ represents the true global minima. When we apply the ADED algorithm to this problem, it efficiently explores the search space, adaptively adjusting its parameters like mutation scale factor (F) and crossover-rate (CR) to guide the search.

Table 11. Pseudocode for ADED single objective

```
function ADED(objective_function, bounds, population_size, max_generations):
    initialize_population()
    initialize_parameters()

    best_solutions = [None] * population_size
    best_fitness_history = []

    for generation in range(max_generations):
        new_population = []

        for i, individual in enumerate(population):
            adapt_parameters()
            neighbors = dynamic_neighborhood(i)
            trial_solution = generate_trial_solution(individual, neighbors)
            trial_fitness = evaluate_fitness(trial_solution)
            selected_individual = crowding_selection(individual, trial_solution, neighbors)
            update_neighborhoods(i, neighbors, trial_fitness)

            new_population.append(selected_individual)

            if best_solutions[i] is None or trial_fitness < evaluate_fitness(best_solutions[i]):
                best_solutions[i] = trial_solution

        population = new_population

        # Record the best fitness for each individual in this generation
        best_fitness_history.append([evaluate_fitness(ind) for ind in best_solutions])

    return best_solutions, best_fitness_history

function initialize_population():
    // Create an initial population of individuals within the specified bounds
    population = [random_uniform(low, high, len(bounds)) for _ in range(population_size)]

function initialize_parameters():
    // Initialize the parameters F and CR
    F = random_uniform(0.5, 2.0)
    CR = random_uniform(0.1, 0.9)

function adapt_parameters(F, CR, fitness, avg_fitness):
    // Adapt parameters (for simplicity, F and CR remain constant)
    return F, CR
```

```
function dynamic_neighborhood(individual_index):
    // Dynamic neighborhood topology (randomly select neighbors)
    all_individuals = list(range(population_size))
    all_individuals.remove(individual_index)  // Exclude self from potential neighbors
    neighborhood_size = min(len(all_individuals), len(neighborhoods[individual_index]))
    return random_choice(all_individuals, neighborhood_size, replace=False)

function generate_trial_solution(individual, neighbors):
    // Generate a trial solution using differential evolution
    r1, r2, r3 = random_choice(len(population), 3, replace=False)
    r1, r2, r3 = population[r1], population[r2], population[r3]
    return individual + F * (r1 - individual) + F * (r2 - r3)

function evaluate_fitness(individual):
    // Evaluate the fitness of an individual using the objective function
    return objective_function(individual)

function average_fitness(population):
    // Calculate the average fitness of the population
    return mean([evaluate_fitness(individual) for individual in population])

function crowding_selection(individual1, individual2, neighbors):
    // Select the individual with better fitness
    if evaluate_fitness(individual1) < evaluate_fitness(individual2):
        return individual1
    else:
        return individual2

function update_neighborhoods(individual_index, neighbors, trial_fitness):
    // Update neighborhood based on trial fitness
    for neighbor_index in neighbors:
        if trial_fitness < evaluate_fitness(neighborhoods[individual_index][neighbor_index]):
            neighborhoods[individual_index][neighbor_index] = trial_fitness
```

ADED increases the performance of the standard DE algorithm by allowing it to explore different regions of the search space. It adapts effectively to the fitness landscape's properties by using adaptive mutation, dynamic neighborhood updates, and local search. Performance and adaptability are influenced by dynamic neighborhood updates, mutation and crossover-rates, and their adaptive nature. These features collectively aim to enhance the algorithm's performance in handling complex and non-convex optimization problems, making it suitable for applications like supply chain analytics. Table 12 presents these changes.

Table 12. ADED vs Classic DE.

| Features | Classic DE | ADED |
|---|---|---|
| $Mutation\ rate\ (F) = initial_{mutation_{rate}} * \left(1 - \frac{generation}{\max_generations}\right)$ | Fixed mutation rate throughout the optimization process. | Adaptive mutation rate dynamically adjusted based on the generation progress. |
| $Crossover\ rate\ (CR) = initial_crossover_rate * \left(\frac{generation}{\max_generations}\right)$ | Fixed crossover-rate throughout the optimization process. | Adaptive crossover-rate dynamically adjusted based on the generation progress. |
| $Mutation\ Strategy\ (v_i) = x_{r_1} + F * (X_{r_2} - X_{r_3})$ $r_1, r_2, r_3$ are distinct random indices representing different solutions in the population, F is the adaptive mutation factor. | Limited set of fixed strategies. | Diverse set of mutation strategies introduced to enhance exploration. |
| $Crossover\ operation\ (u_i[j]) = \begin{cases} v_i[j] & if\ j\ is\ random\ (0,1) < CR \\ x_i[j] & otherwise \end{cases}$ | Fixed crossover operation. | Adaptive crossover operation combining trial solution with original solution. |
| Local Search | Not typically incorporated. | Local search applied after generating trial solution for refinement. |
| Convergence Threshold | Convergence criteria based on fixed number of generations. | Early stopping criteria based on fitness values stagnation for specified generations. |
| Adaptability to Landscape Complexity | Fixed parameter values may lead to premature or slow convergence in challenging landscapes. | Dynamic parameter adjustment and diverse mutation strategies enhance adaptability to complex landscapes. |
| Termination Criteria | Fixed termination criteria such as maximum generations. | Early stopping with better chance of reaching global optimum due to adaptability. |
| Performance in Non-Convex Landscapes | Struggle to escape local optima due to lack of adaptability in parameter settings. | Excels in complex landscapes with multiple local optima, adaptability allows effective exploration and exploitation. |

To this end, we believe that the algorithm has the potential to solve supply chain optimization difficulties and other optimization problems across a range of disciplines by dynamically altering its parameters, employing multiple mutation tactics, and integrating local search. More tests have been performed to assess the performance of ADED, followed by empirical discussions in Sections 3 and 4, respectively.

## 3. Single objective: Benchmark functions

Benchmark functions are crucial for evaluating and developing optimization methods, providing a controlled environment for assessing algorithm performance. Test functions are classified based on surface shape, such as multiple local minima, plate-shaped, valley-shaped, and others, representing increasing difficulty in optimizing test functions. The simulation experiment is programmed with the software of Google Cloud using Jupyter Notebook and Python v3.10.12, and the experiment is configured in Windows 11.0 with Intel(R) Core (TM) i5-4570 CPU @ 3.20GHz, 16.0 GB.

### 3.1 Multiple local optima

In this kind of scenario, the best feasible solution to the problem is the global optimum. Details of the benchmark test conducted in such a scenario are explained in Table 13. All the tests are performed using two dimensions: 300 population sizes, 200 generations, and 30 runs. The number of runs determines how many iterations each algorithm will perform before stopping.

Table 13. Benchmark test functions for many local optima.

| Sl no | Function name | Functions | Ranges |
|---|---|---|---|
| 1 | Ackley (Ackley, 1987) | $-20\, exp\left(-0.2\sqrt{\frac{1}{d}\sum_{i=1}^{d} x_i^2}\right) - exp\left(\frac{1}{d}\sum_{i=1}^{d} cos(2\pi x_i)\right) + 20 + exp(1)$ | $x_i \in [-32.768, 32.768]$, for all i = 1, …, d, although it may also be restricted to a smaller domain. |
| 2 | Bukin N.6 (Bukin, 2007) | $100\sqrt{x_2 - 0.01x_1^2} + 0.01(x_1 + 10)$ | $x_1 \in [-15, -5]$, $x_2 \in [-3, -3]$. Here global minima $f_{bukin}(x^*) = 0.00$ |
| 3 | Rastrigin (Rastrigin, 1974) | $10n + \sum_{i=1}^{n} [x_i^2 - 10Cos(2\pi x_i)]$ | $x_i \in [-5.12, 5.12], for\ all\ i = 1, \ldots, n.$ Here global minima$f_{rastrigin}(0, \ldots, 0) = 0.00$ |
| 4 | Cross-in-tray (Jamil & Yang, 2013) | $f_{crossintray}(x) = -0.0001\left(\left|sin(x_1)sin(x_2)exp\left(\left|100 - \frac{\sqrt{x_1^2+x_2^2}}{\pi}\right|\right)\right|+1\right)^{0.1}$ | $min = \begin{cases} f_{crossintray}(1.35,-1.35) = 2.06262 \\ f_{crossintray}(1.35,1.35) = 2.06262 \\ f_{crossintray}(-1.35,1.35) = 2.06262 \\ f_{crossintray}(-1.35,-1.35) = 2.06262 \end{cases}$, with search domain $-10 \le x, y \le 10$ |
| 5 | Levy N.13 | $sin^2(3\pi x_1) + \sum_i^{n-1}(x_1-1)^2[1+sin^2(3\pi x_2)] + (x_2-1)^2[1+sin^2(2\pi x_2)]$ | $x_i \in [-10,10],\ for\ all\ i = 1,\dots,n.$ Here global minima $f_{levy}(1,1) = 0.00$ |
| 6 | Egg-holder | $-(x_2+47)sin\left(\sqrt{\left\|x_2+\frac{x_1}{2}+47\right\|}\right) - x_i sin\left(\sqrt{\|x_1-(x_2+47)\|}\right)$ | $x_i \in [-512, 404.2319], for\ all\ i = 1,2.$ Here global minima $f_{eggholder}(x^*) = -959.6407$ |
| 7 | Schaffer function N. 2 | $f_{schaffer}(x) = 0.5 + \frac{sin^2(x_1^2 - x_2^2) - 0.5}{[1+0.001(x_1^2 - x_2^2)]^2}$ | $x_i \in [-100,100], for\ all\ i = 1,2.$ Here global minima $f_{schaffer}(x^*) = 0.00$ |
| 8 | Schwefel | $418.9829 * 2 - (x * sin(\sqrt{\|x\|})) + -(y * sin(\sqrt{\|y\|}))$ | $x_i \in [-500,500], for\ all\ i = 1,\dots,n.$ Here global minima $f_{schwefel}(x^*) = 0.00$ |
| 9 | Shubert 3.0 | $\left(\sum_{i=1}^{5} i\ cos((i+1)x_1+i)\right)\left(\sum_{i=1}^{5} i\ cos((i+1)x_2+i)\right)$ | $x_i \in [-10,10], for\ all\ i = 1,2.$ Here global minima $f_{shubert}(x^*) = -186.7309$ |
| 10 | Drop-wave | $-\frac{1+cos\left(12\sqrt{x_1^2+x_2^2}\right)}{0.5(x_1^2+x_2^2)+2}$ | $x_i \in [-5.12, 5.12], for\ all\ i = 1,2.$ Here global minima $f_{dropwave}(x^*) = -1$ |
| 11 | Himmelblau (Himmelblau, 2018) | $(x^2+y-11)^2 + (x+y^2-7)^2$ | $min = \begin{cases} f_{himmelnlau}(3.00,2.00) = 0 \\ f_{himmelnlau}(-2.80,3.13) = 0 \\ f_{himmelnlau}(-3.78,-3,28) = 0' \\ f_{himmelnlau}(3.58,-1.84) = 0 \end{cases}$ with search domain $-5 \le x, y \le 5$ |

Table 14 presents the statistical comparison. Significant differences in mean fitness values and low p-values suggest that ADED outperforms DE on several benchmark functions. Besides t-statistics and p-values, the mean and variability ($\sigma$) are reported. In all the test functions, ADED achieved the global minimum with just 10 runs, which is promising. A grid-based parameter setting can be employed here to optimize the performance of algorithms for specific problems. ADED exhibits characteristics of hill climbing as it iteratively explores the solution space, generates candidate solutions, and aims to improve the current solutions based on their fitness values.

Table 14. ADED with many local optima benchmark test report.

| Sl no | Benchmark test function | | Bounds | $\mu$ | $\sigma$ | t-stats | p-value | Comments |
|---|---|---|---|---|---|---|---|---|
| 1 | Rastrigin | Classic DE | (-5.12, 5.12) | 8.429 | 3.733 | -6.773 | 0.000*** | Nonconvex, multimodal, shifted, and separable |
| | | ADED | | 0.000 | 0.000 | | | |
| 2 | Ackley | Classic DE | (-32.768, 32.768) | 11.460 | 3.913 | -8.784 | 0.000*** | Nonconvex, multimodal |
| | | ADED | | 0.000 | 0.000 | | | |
| 3 | Cross in tray | Classic DE | (-10, 10) | -1.935 | 0.065 | -3.905 | 0.001*** | Nonconvex, multimodal |
| | | ADED | | -2.062 | 0.000 | | | |
| 4 | Egg holder | Classic DE | (-512, 512) | -1446.82 | 372.62 | 3.922 | 0.000*** | Nonconvex, multimodal |
| | | ADED | | -959.640 | 0.000 | | | |
| 5 | Drop-wave | Classic DE | (-5.12, 5.12) | -0.974 | 0.031 | -2.449 | 0.024** | Nonconvex, unimodal |
| | | ADED | | -1.000 | 0.000 | | | |
| 6 | Levy | Classic DE | (-10, 10) | 4.379 | 3.365 | -3.904 | 0.001*** | Nonconvex, multimodal |
| | | ADED | | 0.000 | 0.000 | | | |
| 7 | Schwefel | Classic DE | (-500, 500) | -0.0732 | 0.223 | 2.930 | 0.008*** | |
| | | ADED | | -0.995 | 0.000 | | | Nonconvex, multimodal, shifted |
| 8 | Schaffer N2 | Classic DE | (-100, 100) | -0.073 | 0.223 | -12.380 | 0.000*** | Nonconvex, multimodal |
| | | ADED | | -0.995 | 0.000 | | | |
| 9 | Bukin N6 | Classic DE | (-15, -5) | 25.450 | 5.211 | -14.651 | 0.000*** | Nonconvex, unimodal |
| | | ADED | | 0.000 | 0.000 | | | |
| 10 | Shubert 3.0 | Classic DE | (-10, 10) | -3292.422 | 624.517 | -3.044 | 0.010*** | Nonconvex, multimodal |
| | | ADED | | -3840.000 | 0.000 | | | |
| 11 | Himmelblau | Classic DE | (-5, 5) | 0.000 | 0.000 | 2.627 | 0.017*** | Nonconvex, multimodal |
| | | ADED | | 0.000 | 0.000 | | | |

### 3.3 Plate-Shaped

It is an extension of optimal control theory, with the minimizing parameter being simply the domain in which the problem is described. There are frequently several equal or nearly similar solutions for plate-shaped functions. Since the objective function values of these solutions are quite similar, it is challenging for optimization algorithms to discriminate between them and choose the optimal one. Table 15 displays the test function.

Table 15. Benchmark test functions for plate shaped.

| Sl no | Function name | Functions | Ranges |
|---|---|---|---|
| 12 | Booth | $(x_1 + 2x_2 - 7)^2 + (2x_1 + x_2 - 5)^2$ | $x_i \in [-10, 10], for\ all\ i = 1, 2$. Here global minima $f_{booth}(x^*) = 0.00$ |
| 13 | Matyas | $0.26(x_1^2 + x_2^2) - 0.48x_1x_2$ | $x_i \in [-10, 10], for\ all\ i = 1, 2$. Here global minima $f_{matyas}(0, 0) = 0.00$ |
| 14 | McCormick | $sin(x_1 + x_2) + (x_1 + x_2)^2 - 1.5x_1 + 2.5x_2 + 1$ | $x_1 \in [-1.5, 4], x_2 \in [-3, 4]$. Here global minima $f_{mccormick}(-0.54719, -1.54719) = -1.9133$ |

Table 16 presents the performance report, which clearly shows the superiority of ADED over DE in all categories. Here too, all the tests are performed using the same configurations. We see that ADED achieves the global optima in all three tests with just 10 runs.

Table 16. ADED with plate shape benchmark test report.

| Sl no | Benchmark function | | Bound | $\mu$ | $\sigma$ | t-stats | p-value | Comments |
|---|---|---|---|---|---|---|---|---|
| 12 | McCormick | Classic DE | (-15, 4) | -1.913 | 2.230 | 4.113 | 0.000*** | Convex, unimodal, separable |
| | | ADED | | -4.972 | 0.000 | | | |
| 13 | Matyas | Classic DE | (-10, 10) | 0.585 | 0.757 | -2.318 | 0.004*** | Nonconvex, unimodal, separable |
| | | ADED | | 0.000 | 0.000 | | | |
| 14 | Booth | Classic DE | (-10, 10) | 10.417 | 0.000 | -6.309 | 0.000*** | Convex, unimodal, separable |
| | | ADED | | 0.000 | 0.000 | | | |

### 3.3 Valley-Shaped

Long, narrow valleys with several local optima are common in valley-shaped functions. Optimization algorithms may struggle to avoid these valleys, resulting in convergence to suboptimal solutions. Table 17 displays the test functions.

Table 17. Benchmark test functions for valley shaped.

| Sl no | Function name | Functions | Ranges |
|---|---|---|---|
| 15 | Three-Hump Camel | $2x_1^2 - 1.05x_1^4 + \frac{x_1^6}{6} + x_1x_2 + x_2^2$ | $x_i \in [-5,5], for\ all\ i = 1,2$. Here global minima $f_{threehumpcamel}(x^*) = 0$ |
| 16 | Six-Hump Camel | $\left(4 - 2.1x_1^2 + \frac{x_1^4}{3}\right)x_1^2 + x_1x_2 + (-4 + 4x_2^2)x_2^2$ | $x_1 \in [-3,3],\ x_2 \in [-2,2]$. Here global minima $f_{sixhumpcamel}(x^*) = 1.0316$ |
| 17 | Rosenbrock | $\sum_{i}^{n-1}[100(x_{i+1} - x_i^2)^2 + (x_i - 1)^2]$ | $min = \begin{cases} n = 2 \rightarrow f_{rosenbrock}(1,1) = 0 \\ n = 3 \rightarrow f_{rosenbrock}(1,1,1) = 0 \\ n > 3 \rightarrow f_{rosenbrock}(1,\ldots,1) = 0 \end{cases}$, with search domain $-\infty \leq x_i \leq \infty, 1 \leq i \leq n$ |
| 18 | Dixon-Price | $(x_i - 1)^2 + \sum_{i=2}^{n}[i(2x_i^2 - x_{i-1})^2]$ | $x_i \in [-10,10], for\ all\ i = 1,\ldots,n$. Here global minima $f_{dixonprice}(x^*) = 0$ |

Table 18 displays the report with the clear superiority of ADED over DE in all categories. The Rosenbrock function is unimodal. Its minimum is tucked away in a valley with a flat bottom and a banana-shaped shape. Here, the algorithm required more than 100 runs to succeed.

Table 18. ADED with valley-shape benchmark test report

| Sl no | Benchmark function | | Bounds | $\mu$ | $\sigma$ | t-stats | p-value | Comments |
|---|---|---|---|---|---|---|---|---|
| 15 | Rosenbrock | Classic DE | (-5, 10) | 50.138 | 53.356 | -2.819 | 0.011*** | Nonconvex, unimodal |
| | | ADED | | 0.000 | 0.000 | | | |
| 16 | Three-hump camel | Classic DE | (-5, 5) | 0.784 | 0.481 | -4.883 | 0.000*** | Nonconvex, unimodal, shifted, and separable |
| | | ADED | | 0.000 | 0.001 | | | |
| 17 | Six-hump camel | Classic DE | (-3, 3) | -0.759 | 0.167 | -4.865 | 0.000*** | Nonconvex, multimodal, shifted |
| | | ADED | | -1.031 | 0.000 | | | |
| 18 | Dixon price | Classic DE | (-10, 10) | 65.246 | 50.986 | -3.839 | 0.001*** | Nonconvex, unimodal, shifted |
| | | ADED | | 0.000 | 0.000 | | | |

## 3.4 Other

Table 19 displays the functions that provide a standardized way to evaluate optimization algorithms' performance across different problem landscapes and complexities.

Table 19. Benchmark test functions for other functions.

| Sl no | Function name | Functions | Ranges |
|---|---|---|---|
| 19 | Beale | $(1.5 - x_1 + x_1x_2)^2 + (2.25 - x_1 + x_1x_2^2)^2 + (2.625 - x_1 + x_1x_2^3)^2$ | $x_i \in [-4.5, 4.5], for\ all\ i = 1, 2$. Here global minima $f_{beale}(3, 0) = 0$ |
| 20 | Goldstein | $1 + (x_1 + x_2 + 1)^2 + (19 - 14x_1 + 3x_1^2 - 14x_2 + 6x_1x_2 + 3x_2^2)] * [(30 + (2x_1 + 3x_2)^2(18 - 32x_1 + 12x_1^2 + 48x_2 - 36x_1x_2 + 27x_2^2)$ | $x_i \in [-2, 2], for\ all\ i = 1, 2$. Here global minima $f_{goldsteinprice}(0, -1) = 3$ |
| 21 | Forrester (Forrester et al. 2008) | $(6x - 2)^2 sin(12x - 4)$ | x ∈ [0,1] |
| 22 | DeVilliersGlasser02 Gavana (2016 and subsequently Layeb (2022) found that DeVilliersGlasser02 is harder to solve than others. | $(2x_1 - 3x_2)^2 + 18x_1 - 32x_2 + 12x_1^2 + 48x_2 + 27x_2^2$ | $x \in [1, 60]\ for\ 1, \ldots, n$. Here the global optimal $f_{devilliersglasser02}(x^*) = 0$ |

Table 20 presents the consolidated report. Here too, ADED outperforms DE in most of the categories except Beale, where, although the global minima are achieved by both DE and ADED, the output is not statistically significant. It suggests that the observed variations in performance between the two groups (in this case, ADED and DE) could be attributable to random variability or by chance rather than a systematic and meaningful difference.

Table 20. ADED with other benchmark test reports.

| Sl no | Benchmark function | | Bound | $\mu$ | $\sigma$ | t-stats | p-value | Comments |
|---|---|---|---|---|---|---|---|---|
| 19 | Beale | Classic DE | (-4.5, 4.5) | 0.000 | 0.000 | 1.000 | 0.330 | Nonconvex, multimodal, separable |
| | | ADED | | 0.000 | 0.000 | | | |
| 20 | Goldstein price | Classic DE | (0, -1) | 44.868 | 38.521 | -3.260 | 0.000** | Nonconvex, multimodal |
| | | ADED | | 2.990 | 0.000 | | | |
| 21 | Forrester | Classic DE | (0, 1) | -44.183 | 17.378 | 5.548 | 0.017*** | Nonconvex, unimodal |
| | | ADED | | -12.041 | 0.000 | | | |
| 22 | DeVilliersGlasser02 | Classic DE | (1, 60) | 143.644 | 201.111 | -6.576 | 0.000*** | Nonconvex, multimodal |
| | | ADED | | 47.938 | 142.900 | | | |

To assess the algorithm's performance and applicability for various optimization challenges, we tested it against a variety of benchmark functions. Appendix 1 displays the success rate for various test functions with the number of iterations. We can identify that for some functions, e.g., Shubert, Himmelblau, and Beale, both algorithms struggle to find the global minimum in all the iterations (number of runs).

## 3.5 Mult objective benchmark test functions

The main justification for a problem's multi-objective formulation is the impossibility of finding a single solution that perfectly balances every objective. As a result, an algorithm is very useful if it provides many potential solutions that are on or close to the Pareto-optimal front (Adeyemo & Otieno, 2009). Here, the main idea is to enhance its ability to converge to Pareto optimal solutions in a multi-objective optimization situation by combining the global exploration capabilities of differential evolution with a self-adaptive local search mechanism. The algorithm's ability to self-adapt enables it to dynamically modify its parameters as it goes through the optimization process, increasing its robustness and efficiency. Multi-objective optimization aims to find alternatives as diverse as possible inside the resulting non-dominated front and find solutions as close to the Pareto-optimal solutions as feasible. Thus, there are two objectives – generational distance ($gd$) and diversity metrics ($\Delta$) and if both these objectives are met, a good multi-objective evolutionary algorithm will be identified (Deb, 2001).

Table 37 displays the pseudocode for ADED multi-objective with scalarization and pareto dominance functions.

Table 21. ADED multi-objective additional functions

```
Function multi_objective_function(x):
    obj1 = sin(x[0]) + cos(x[1])
    obj2 = exp(-(x[0] - 5)^2 - (x[1] - 5)^2)
    Return [obj1, obj2]
End Function

Function scalarization_function(objective_values):
    weights = [0.5, 0.5]
    Return DotProduct(objective_values, weights)
End Function

Function pareto_dominance(obj_values1, obj_values2):
    For each index i from 0 to Length(obj_values1) - 1:
        If obj_values1[i] > obj_values2[i]:
            Return False  # obj_values1 is not dominating obj_values2
    End For

    If any value in [val1 < val2 for each (val1, val2) pair in
Zip(obj_values1, obj_values2)]:
        Return True
    Else:
        Return False
obj_values2
End Function
```

- The generational distance (GD) metric quantifies the average distance of solutions in set Q from the Pareto-optimal solutions represented by set $P^*$. Mathematically, GD is defined as follows:

$$gd = \sqrt{\frac{1}{|P^*|}\sum_{p^* \in P^*} d^2(p^*, Q)}$$

  Here, $|P^*|$ is the cardinality (number of elements) of the Pareto-optimal set, $p^*$ represents an individual solution in the Pareto-optimal set $d(p^*, Q)$ is the Euclidean distance between the solution and its nearest neighbor in set Q.

- The spread measure ($\Delta$) assesses the dispersion achieved using the non-dominated solutions and is defined as:

$$\Delta = \frac{d_f + d_l + \sum_{i=1}^{|Q|-1}(d_i - \bar{d})}{d_f + d_l + (|Q| - 1)d}$$

where, $d_i$ is the Euclidean distance (measured in the objective space) between consecutive solutions in the obtained non-dominated front Q and $\bar{d}$ is the average of these distances. $d_f$ and $d_l$ are the Euclidean distances between the extreme solutions of the Pareto front $P^*$ and the boundary solution of the obtained front $Q$. $\Delta$ provides insights into how well-distributed the solutions in Q are in the objective space. It considers both the average distances between consecutive solutions and the distances from the extreme solutions of the true Pareto front, contributing to a comprehensive evaluation of the spread of non-dominated solutions.

Table 22. Pseudocode for ADED multi-objective implementation.

```
Function ADED_multi_objective(objective_function, bounds, population_size, max_generations, stagnation_limit=10,
                initial_mutation_rate=0.5, initial_crossover_rate=0.9):
  # Initialize the population
  population = initialize_population(population_size, bounds)

  # Initialize mutation and crossover rates
  F = adaptive_mutation_rate(0, max_generations, initial_mutation_rate)
  CR = adaptive_crossover_rate(0, max_generations, initial_crossover_rate)

  # Initialize best solution and history
  best_solution = None
  best_fitness_history = []

  # Main loop for generations
  for generation in range(max_generations):
    new_population = []

    # Loop through each individual in the population
    for i, individual in enumerate(population):
      # Update mutation and crossover rates
      F = adaptive_mutation_rate(generation, max_generations, initial_mutation_rate)
      CR = adaptive_crossover_rate(generation, max_generations, initial_crossover_rate)

      # Randomly select two distinct neighbors
      neighbors = list(range(len(population)))
      neighbor1 = population[np.random.choice(neighbors)]
      neighbor2 = population [np.random.choice(neighbors, exclude=neighbor1)]

      # Generate trial solution using DE mutation strategy
      trial_solution = individual + F * (neighbor1 - individual) + F * (neighbor2 - individual)

      # Apply local search to the trial solution
      trial_solution = local_search(trial_solution, objective_function, bounds)

      # Evaluate fitness of the trial solution
      trial_fitness = objective_function(trial_solution)

      # Pareto dominance check with the new population
      dominated = any(pareto_dominance(trial_fitness, objective_function(ind)) for ind in new_population)

      # If not dominated, add to the new population
      if not dominated:
        new_population.append(trial_solution)

      # Update the best solution based on Pareto dominance
      if best_solution is None or pareto_dominance(trial_fitness, objective_function(best_solution)):
        best_solution = trial_solution

    # Update the population with the new one
    population = new_population

    # Update the best fitness history
    best_fitness_history.append(objective_function(best_solution))

    # Check for convergence
    if has_converged(best_fitness_history, stagnation_limit):
      break
```

```
    # Return the best solution found
    Return best_solution
End Function
```

Table 23 displays the main changes between the multi-objective version and the single-objective version of the ADED algorithm in how solutions are evaluated, compared, and selected for the next generation. These changes reflect the inherent differences in handling multiple objectives, where the optimization goal is to find a set of solutions that represents a trade-off between conflicting objectives, rather than converging to a single optimal solution.

Table 23. ADED single vs multi objective.

| **Single objective** | **Multi objective** |
|---|---|
| The fitness of an individual is determined by a single objective function value. | The fitness of an individual is represented by a vector of objective values. Each objective contributes to the overall fitness. |
| Individuals are compared based on their fitness values. The comparison is straightforward – a lower fitness value is considered better. | Individuals are compared using Pareto dominance. An individual A is considered better than individual B if it is at least as good as B in all objectives and strictly better in at least one objective. |
| An individual is selected for the next generation if its trial solution has a better fitness than the current individual. | An individual is selected for the next generation only if it is not dominated by any individual in the new population. |
| The best solution is updated based on the trial solution's fitness if the trial solution is better. | The best solution is updated based on Pareto dominance. The trial solution becomes the best solution only if it is non-dominated. |

Here, the benchmark test for multi-objective is tested with ZDT test suite. This popular test suite was designed for two-objective issues and was named for its creators, Zitzler, Deb, and Thiele (Zitzle et al., 2000).

Table 24. ZDT test suite for multi-objective functions.

| **Sl no** | **Name** | **Functions** | **Ranges** |
|---|---|---|---|
| 23 | ZDT1 | $objective\ a\ (f_1): x_1$ <br> $objective\ b\ (f_2): g * \left(1 - \sqrt{\frac{x_1}{g}}\right), where\ g = 1 + 9 * \left(\frac{\sum_{i=2}^{n} x_i}{n-1}\right)$ | $x_1 \in [0,1]$) and $x_1, x_2, x_3 \dots\ x_n$ (real numbers in the range [0.1]). The Pareto-optimal front for ZDT1 is a concave curve. The optimization goal is to find a set of solutions on the Pareto front that represents a trade-off between minimizing $f_1$ and $f_2$. |
| 24 | ZDT2 | $f_1(x) = x_1$ <br> $g(x) = 1 + 9 * \left(\frac{\sum_{i=2}^{n} x_i}{n-1}\right)$ <br> $f_2(x) = g(x)\left(1 - \left(\frac{f_1(x)}{g(x)}\right)^2\right)$ | $x_1 \in [0,1]$) |
| 25 | DLTZ1 | $f_1(x) = \frac{1}{2} * (1 + g(x)) * x_1 * x_2 \dots * x_{M-1}$ <br> $f_2(x) = \frac{1}{2} * (1 + g(x)) * x_1 * x_2 \dots * (1 - x_{M-1})$ <br> $f_3(x) = \frac{1}{2} * (1 + g(x)) * x_1 * x_2 \dots * (1 - x_{M-2}) * x_{M-1}$ <br> $\vdots$ <br> $f_M(x) = \frac{1}{2} * (1 + g(x)) * (1 - x_1)$ | $x_i \in$ [0, 1], i = 1, . . ., M − 1 the function corresponds to the surface of a sphere in the first orthant of M dimensional space. $g(x)$ is the linking function that depends on the decision variable $x_i$ and is defined as: $g(x) = 100\left\{(M-1) + \sum_{i=1}^{M-1}\left(x_i - \frac{1}{2}\right)^2 - \cos\left(20\pi\left(x_i - \frac{1}{2}\right)\right)\right\}$ The objective is to minimize all M objectives simultaneously. |

Table 25. Multi-objective problems – test statistics

| Benchmark test | Generational distance (GD) | Diversity metrics | Comments |
|---|---|---|---|
| ZDT1 | 0.180276 | $1.64\times10^{-9}$ | GD of 0.180 suggests that the algorithm is producing solutions that are quite close to the true Pareto front. Δ is very close to zero, suggesting that the solutions in the obtained Pareto front are very similar to each other. |
| ZDT2 | 0.272168 | 0.030310 | GD value of 0.396 suggests that the obtained Pareto front is, on average, 0.396 units away from the true Pareto front. Δ of 0.0303 indicates that the solutions in the obtained Pareto front are relatively diverse, which is a positive aspect. |
| DLTZ1 | 0.229283 | 0.837994 | on average, produced solutions are relatively close to the true Pareto front (low GD). The obtained Pareto front is somewhat diverse (moderate diversity metric), but there may be room for improvement to achieve a broader spread of solutions. |

## 4. Empirical results & discussions

Comparative analysis is performed with the Classic DE and the popular variants of DE e.g., composite DE (CoDE) (Wang et al., 2011), JADE (Adaptive DE with External Archive) (Zhang & Sanderson, 2009), jDE (Self-Adaptive DE) (Brest et al., 2009) and SaDE (Self-Adaptive DE) (Qin & Suganthan, 2005). Fig. 14 and Fig. 15 display a comparison of the convergence behavior of different optimization algorithms on the convex and nonconvex objective functions over multiple (30) runs. These plots show how quickly and effectively each algorithm converges to a solution in the given search space. The x-axis represents the generation number, and the y-axis represents the best fitness value found by the algorithm in each generation.

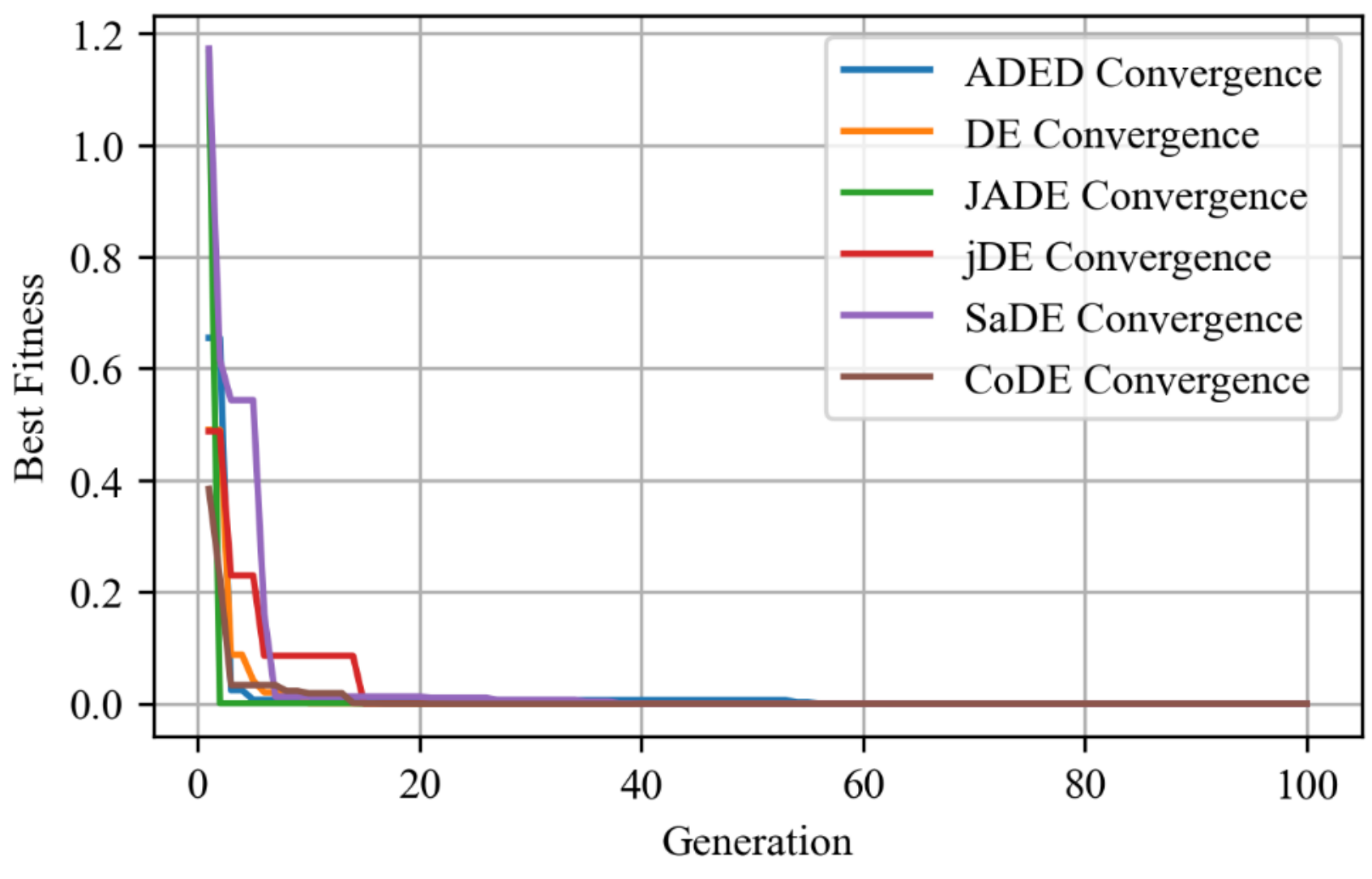

Fig. 14. Convergence history {$objective\ function\ f(x) = \sum x^2$, average 30 runs}.

Fig. 14 shows that, both ADED and JADE exhibit fast adaptation and quick convergence with quadratic convex objective function. Both shows sharp drop along the best fitness line towards the global minimum. Other algorithms reach the global minimum after initial struggle and require more iterations. They show complex convergence behavior before reaching to optimal line.

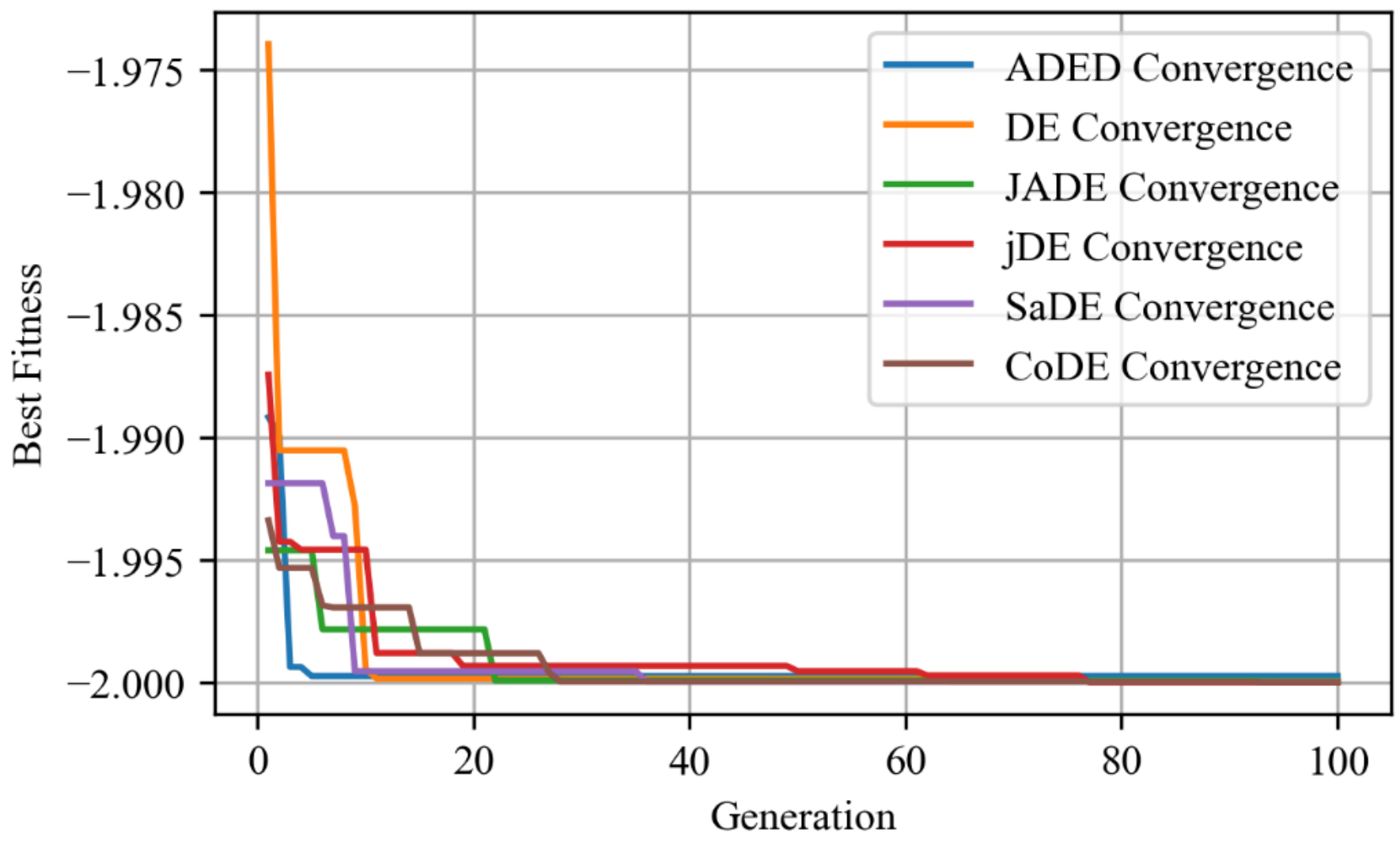

Fig. 15. Convergence history {$objective\ function\ f(x) = sin(x_1) + cos(x_2)$, average 30 runs}.

Fig. 15 displays the convergence behavior of the algorithms (ADED, DE, JADE, jDE, SaDE, and CoDE) over generations. ADED starts at the somewhat middle point (-1.990) on the best fitness line and exhibits a sharp drop in the fitness value, reaching the optimal value of -2.00. This indicates that ADED is capable of quickly adapting its strategy and exploring the search space effectively to find the global minimum. The convergence pattern of other algorithms (DE, JADE, jDE, SaDE, and CoDE) show complex behavior which require several iterations before reaching the optimal line of global minimum.

## 4.1 Analytical measurements

The performance of evolution algorithm depends on parameter selection (mutation and crossover strategies). According to the DE theory, a target vector of $x_i^t$ serves as the foundation for the construction of a mutant or donor vector, known as $y_i^t$, through a mutation process. Though the default is $best1bin$, however, ADED is evaluated with 14 possible mutation and crossover strategies which are obtained by combining binomial and exponential crossover systems with the commonly used differential mutation procedures displayed in Table 26. These variants control how the diversity of the population is maintained and how information from different individuals is used to explore the search space.

Table 26. Mutation and crossover strategies.

| Mutation strategy | Description | Variant | Description |
|---|---|---|---|
| $rand/1$ | A mutant vector is generated by adding a scaled difference vector to a randomly chosen vector from the population. | $X_{r_1^i,G} + F\left(X_{r_2^i,G} - X_{r_3^i,G}\right)$ | $r_1$ is the population index chosen as the base vector; $r_2$, $r_3$, $r_4$, and $r_5$ are the population indices randomly chosen to create the mutant vector; and $r_1$, $r_2$, $r_3$, $r_4$, $r_5$ $\in$ [1, population size] and $r_1 \neq r_2 \neq r_3 \neq r_4 \neq r_5 \neq$ i; suggest that the best individual solution within the DE population is chosen as the target vector; F is a scaling factor with a value in the interval [0,1]. Selecting the right value for F is essential to achieving the right balance between exploration and exploitation searches and prevent early or slow convergence. |
| $best/1$ | The mutant vector is created by adding a scaled difference vector to the best vector in the population. | $X_{best,G} + F\left(X_{r_1^i,G} - X_{r_2^i,G}\right)$ | |
| $rand/2$ | Two randomly chosen vectors contribute to the mutant vector. | $X_{r_1^i,G} + F\left(X_{r_2^i,G} - X_{r_3^i,G} + X_{r_4^i,G} - X_{r_5^i,G}\right)$ | |
| $best/2$ | The best vector and another randomly chosen vector contribute to the mutant vector. | $X_{best,G} + F\left(X_{r_1^i,G} - X_{i,G} + X_{r_3^i,G} - X_{r_4^i,G}\right)$ | |
| $current-to-rand/1$ | Combination of the current vector and a randomly chosen vector. | $X_{i,g} + K\left(X_{r_3^i,G} - X_{i,g}\right) + F\left(X_{r_1^i,G} - X_{r_2^i,G}\right)$ | |
| $current-to-best/1$ | Combination of the current vector and the best vector. | $X_{i,g}+K(X_{best,G}-X_{i,g})+F\left(X_{r_1^i,G}-X_{r_2^i,G}\right)$ | |

The convergence nature of ADED is investigated using Quality Measure ($Q_{measure}$). It compares how well various objective functions converge. The $Q_{measure}$ combines the convergence speed and the probability of convergence for a given variant. The convergence speed ($C_s$) is calculated as the minimum fitness value achieved in the last generation of each run. The minimum is taken to represent the fastest convergence. The probability of convergence is the ratio of successful runs (where the algorithm converges) to the total number of runs. The convergence measure is the total number of evaluations divided by the number of successful runs. It represents the efficiency of the algorithm in achieving convergence. The $Q_{measure}$ is then calculated as the convergence measure divided by the probability of convergence.

$$Q_{measure}=\frac{C}{P_{convergence}}$$

- $C = convergence\ measure = \frac{\sum E_j}{R}$
- $\sum E_j = total\ number\ of\ function\ evaluations\ taken\ for\ all\ the\ successful\ runs$
- $R = number\ of\ successful\ runs$
- $P_{convergence} = probability\ of\ convergence$

It checks both convex and non-convex variations for every variant, which is 14 variants * 2 = 28. Each variant is run through 30 times, for a total of 28 * 30 = 840.

## 4.2 Crossover strategy

For every combination of variation and test function, the crossover-rate (CR) was adjusted. Since ADED is a stochastic algorithm, thirty independent runs were made for every combination. For each variant and each function (convex and non-convex), the best fitness values obtained in each run are collected. So, for each variant and function, the algorithm is run 30 times, and the best fitness values are collected for AOV calculation. The AOV is calculated as the average of all these best fitness values. Table 27 displays the variations and CR values for every test function. The table demonstrates the algorithm's performance under various mutation and crossover strategies. The best solution and best fitness values are an indication of how well each variant performs on the optimization task.

Table 27. ADED evaluation with mutation and crossover strategies.

| Sl. no. | Variants | Parameters | | Best solution | Best fitness |
|---|---|---|---|---|---|
| 1 | $rand1bin$ | i | {'F': 0.9, 'CR': 0.5} | [-5.71071350e-14, 2.39724262e-14] | 3.835902087729318e-27 |
| | | ii | {'F': 0.9, 'CR': 0.0} | None | inf |
| 2 | $best1bin$ | iii | {'F': 0.1, 'CR': 0.1} | [ 7.32408627e-08, -1.48320435e-07] | 2.7363175454653217e-14 |
| | | iv | {'F': 0.9, 'CR': 0.7} | [-4.53026614e-16, 3.04820311e-16] | 2.9814853457854587e-31 |
| 3 | $rand2bin$ | v | {'F': 0.3, 'CR': 0.2} | [2.70981361e-11, 6.75006924e-11] | 5.290652449930983e-21 |
| | | vi | {'F': 0.9, 'CR': 0.3} | [-9.40282223e-10, -2.34568235e-09] | 6.386356349321004e-18 |
| 4 | $best2bin$ | vii | {'F': 0.1, 'CR': 0.7} | [-9.98168128e-35, 2.98105740e-34] | 9.883042838814909e-68 |
| | | viii | {'F': 0.9, 'CR': 0.3} | [-9.01349244e-10, -7.20459893e-10] | 1.3314929174430563e-18 |
| 5 | $currenttorand1bin$ | ix | {'F': 0.5, 'CR': 0.4} | [2.98848127e-16, 1.00489327e-15] | 1.0991206778208615e-30 |
| | | x | {'F': 0.9, 'CR': 0.3} | [ 9.77027486e-10, -2.26073344e-10] | 1.005691866201369e-18 |
| 6 | $currenttobest1bin$ | xi | {'F': 0.2, 'CR': 0.8} | [-3.21316123e-34, 6.03926559e-34] | 4.679713400579999e-67 |
| | | xii | {'F': 0.9, 'CR': 0.1} | [ 0.00012376, -0.00021369] | 6.097951666607039e-08 |
| 7 | $randtobest1bin$ | xiii | {'F': 0.1, 'CR': 0.8} | [-6.53633990e-38, -8.28356743e-38] | 1.1134122862155267e-74 |
| | | xiv | {'F': 0.9, 'CR': 0.4} | [-1.77884606e-11, 5.68334766e-12] | 3.487297718495331e-22 |

The competitiveness in solving the benchmark functions is identified by comparing their mean objective function values. The convergence analysis of the variants is carried out by measuring their Convergence Speed, Quality Measure for each variant-test function combination. Table 28 displays information about different variants, their associated performance metrics (AOV, $C_s$, $Q_{measure}$, and average rank), as well as their rankings based on these metrics.

Table 28. Performance Metrics and Rankings of ADED Variants for Optimization.

| Sl no | Variants | Parameters | AOV | AOV Rank | $C_s$ | $C_s$ Rank | $Q_{measure}$ | $Q_{measure}$ Rank | Average Rank |
|---|---|---|---|---|---|---|---|---|---|
| 1 | $rand1bin$ | {'F': 0.9, 'CR': 0.5} | -0.98140 | 13.0 | -0.999945 | 3.0 | 16210778.70 | 7.0 | 7.666667 |
| 2 | $rand1exp$ | {'F': 0.9, 'CR': 0.0} | -0.98768 | 9.0 | -0.999929 | 2.0 | 18062875.28 | 10.0 | 7.000000 |
| 3 | $best1bin$ | {'F': 0.1, 'CR': 0.1} | -0.99754 | 3.0 | -0.999999 | 11.0 | 9143449.64 | 2.0 | 5.333333 |
| 4 | $best1exp$ | {'F': 0.9, 'CR': 0.7} | -0.98879 | 7.0 | -0.999953 | 6.0 | 25895111.11 | 13.0 | 8.666667 |
| 5 | $rand2bin$ | {'F': 0.3, 'CR': 0.2} | -0.99972 | 2.0 | -1.0 | 14.0 | 21363430.83 | 11.0 | 9.000000 |
| 6 | $rand2exp$ | {'F': 0.9, 'CR': 0.3} | -0.98589 | 12.0 | -0.999955 | 7.0 | 16078181.40 | 6.0 | 8.333333 |
| 7 | $best2bin$ | {'F': 0.1, 'CR': 0.7} | -0.99460 | 6.0 | -0.999999 | 10.0 | 10752790.69 | 3.0 | 6.333333 |
| 8 | $best2exp$ | {'F': 0.9, 'CR': 0.3} | -0.98677 | 7.0 | -0.999947 | 6.0 | 27360764.17 | 13.0 | 8.666667 |
| 9 | $currenttorand1bin$ | {'F': 0.5, 'CR': 0.4} | -1.0 | 1.0 | -1.0 | 15.0 | 26086222.22 | 14.0 | 10.0000 |
| 10 | $currenttorand1exp$ | {'F': 0.9, 'CR': 0.3} | -0.97989 | 14.0 | -0.999963 | 8.0 | 17454553.87 | 8.0 | 10.0000 |
| 11 | $currenttobest1bin$ | {'F': 0.2, 'CR': 0.8} | -0.99738 | 4.0 | -1.0 | 13.0 | 13111101.51 | 4.0 | 7.000000 |
| 12 | $currenttobest1exp$ | {'F': 0.9, 'CR': 0.1} | -0.98649 | 11.0 | -0.999948 | 5.0 | 23672641.97 | 12.0 | 9.333333 |
| 13 | $randtobest1bin$ | {'F': 0.1, 'CR': 0.8} | -0.99516 | 5.0 | -0.999999 | 12.0 | 13613615.33 | 5.0 | 7.333333 |
| 14 | $randtobest1exp$ | {'F': 0.9, 'CR': 0.4} | -0.9887 | 8.0 | 0.99997 | 9.0 | 17804044.37 | 9.0 | 8.666667 |

Choosing the most effective mutation scheme depends on the specific characteristics of the optimization problem. Average Rank is a key metric, and a lower average rank indicates better performance across all metrics. The best-performing variant based on average rank is $best1bin$ with an average rank of 5.33, making it the top performer in this analysis. Lower AOV indicates better convergence. The best-performing variant in terms of AOV is $rand2bin$ with a AOV of -0.99972. Lower $C_s$ indicates faster convergence. The best-performing variant in terms of $C_s$ is $rand1bin$ with a $C_s$ of -0.999945. A lower $Q_{measure}$ indicates better overall performance considering both convergence and success rate.

Fig. 21 displays a visual representation of the average rank of each variant based on three performance metrics: AOV, $C_s$, and $Q_{measure}$. The average rank is calculated by taking the average of the ranks for each variant across the three metrics. This provides an overall ranking considering all three performance aspects. The recommended variants are $best1bin$, $best2bin$, and $rand1bin$ based on Average Rank. All these processes involve randomness in the selection of individuals and the crossover process to introduce diversity and enable effective exploration in the optimization process. This way ADED aims to strike a balance between exploration and exploitation by incorporating randomization and a more deterministic approach.

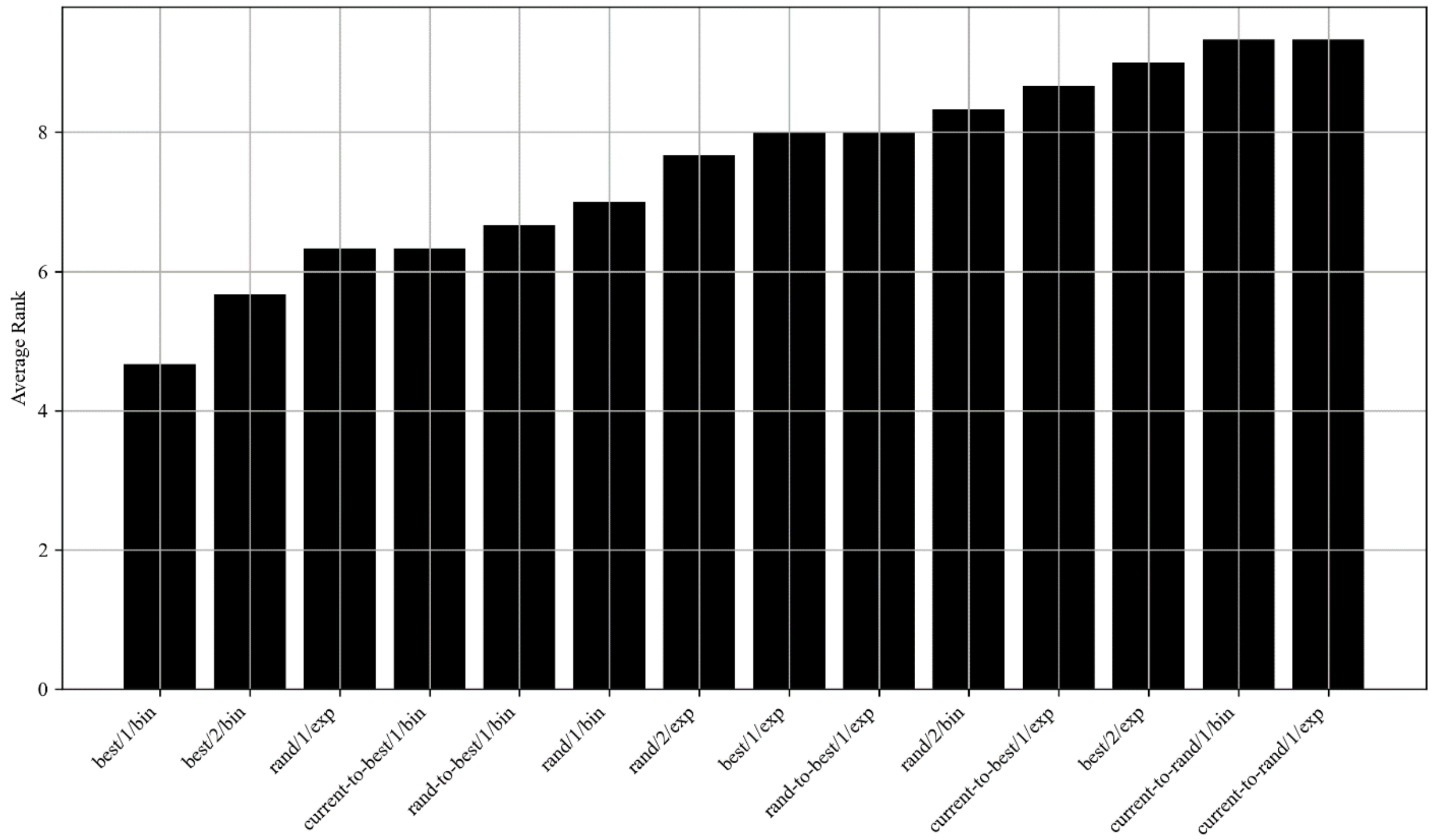

Fig. 16. Average Rank of the variants.

Ahmad et al. conducted an extensive review and found that, most researchers between 2016 – 2021 implemented DE in different areas using the parameter values population size = 100, F, and CR = 0.5 (Ahmad MF et al., 2022). So, we recommend these numbers as the default parameter values.

## 4.1 Practical implications & future directions

The ADED algorithm has shown promising performance across a diverse set of benchmark functions and empirical evaluations, demonstrating its versatility and effectiveness. It outperforms classic DE in terms of convergence speed and solution quality across a wide range of optimization challenges, including functions with many local optima, plate-shaped, valley-shaped, stretched-shaped, and noisy functions. For most of the vital test problems, such as Ackley, Matyas, Booth, Goldstein-Price, Beale, Bukin, Levy, McCormic, Six-Hump Camel, and Three-Hump Camel, the algorithm did exceedingly well. This gives us confidence that ADED is appropriate for supply chain analytics and capable of addressing the complexities and challenges of supply chain management. Its adaptability, performance, and ability to handle various optimization landscapes make it a relevant and promising approach for improving supply chain operations and decision-making. ADED has an extra degree of flexibility that allows it to tackle more complex and nuanced optimization problems in the context of supply chain management because of its capacity to handle multi-objective optimization.

Artificial intelligence related technological innovation has put the application of artificial neural networks (ANN) in the forefront. With the availability of data volume, business often deploy ANN based model to enhance there porductibity (Abualigah et al., 2023; Baioletti et al., 2020). However, there are issues with convergence and subsequently optimization with the existing traditional ANN method. Although ANN are regarded as one of the best generic algorithms for problem solving, they are really a stochastic problem in which model weights are employed and every iteration is restructured with the backpropagation of the error algorithm signal. Since the late 1990s and the beginning of the 2000s, ANN has been using DE-based optimizers (Ilonen et al., 2003; Masters et al., 1997). ADED can improve the optimization process, including adaptive processes, diversification techniques, dynamic neighborhood structures, individual-specific adaptation, crowding selection, and improved convergence monitoring. In contrast to classic DE, ADED adapts the strategy parameters dynamically to the changing optimization issue using adaptive processes. To encourage solution diversity, a greater range of solutions is promoted, and premature convergence is avoided. It also contains a variety of mutation tactics and measures. Because of this, ADED is a good option for the optimization of ANN in high-dimensional, dynamic, and non-convex optimization scenarios. Therefore, the feasibility of using ADED as an optimizer for neural network topologies, including convolutional, recurrent, and mixed neural networks, will be examined in future research.

The adaptive crossover and mutation rates give an edge to ADED at optimizing complex and dynamic settings. Its ability to continuously adjust its settings in response to progress can be used for managing demand volatility in supply chains. Its versatility increases the accuracy of demand predictions by allowing it to effectively explore and use the search field. Inventory levels can be optimized with the use of the algorithm's crossover operation and mutation technique. By analyzing experimental choices based on different mutation strategies and honing them through local search, ADED could help to achieve a balance between meeting demand and decreasing holding costs.

The following is a list of some possible uses for ADED:

a) The adaptability and diversification methods of ADED may make the supply chain more efficient, cost-effective, and productive and assist business managers in addressing the issues of fluctuating demand and inventory control.
b) By making use of its flexible features, ADED can be employed to retain the optimal transit routes in the face of changing variables and respond to changing logistical situations with effectiveness. Cost-effectiveness and prompt delivery—two crucial concerns in the dynamic world of supply chain management and logistics—can be made easier by this flexibility.
c) Adding metrics for adaptation and diversity to ADED significantly enhances supply chain resilience. The algorithm's adaptability allows it to discover solutions in the face of dynamic supply chain circumstances and reduce the impact of unforeseen changes.
d) By adapting to shifting circumstances and variations in demand, ADED may support the optimization of resource allocation. The algorithm's flexibility would help to ensure optimal resource utilization, manage expenses, and effectively meet customer demand.

## Conclusion

The Adaptive Differential Evolution with Diversification (ADED) algorithm introduced in this study stands out as a reliable and adaptive solution for both single- and multi-objective optimization challenges. The key features include adaptive parameter adjustment, the generation of a diverse set of solutions via mutation strategies, safeguards against early convergence, the inclusion of local search mechanisms for refining the solutions, and convergence monitoring. The algorithm is tested on a range of 22 benchmark test functions, with additional ZDT test suites for multi-objective test functions. Strong success rates are noted, confirming ADED's effectiveness in a variety of contexts and compared with the classic DE and its variants, including CoDE, JADE, jDE, and SaDE. Its adaptable parameters enable efficient exploration and quick improvements in fitness. The analysis of 14 mutation and crossover strategies identified two recommended schemes: $best1bin$ and $rand1bin$. The $best1bin$ scheme uses the best individual for mutation and employs binary crossover, making it the top performer, demonstrating better convergence speed, solution quality, and overall performance. ADED's adaptability and dynamic optimization capabilities make it suitable for optimizing inventory levels, production schedules, routing plans, and supplier selections. Future directions include exploring its application as an optimizer for neural network topologies. Additionally, assessing ADED's parallelization capabilities and scalability for large-scale optimization problems could enhance its applicability. Hybridizing ADED with other optimization techniques or problem-specific heuristics remains a promising avenue for developing hybrid algorithms with superior problem-solving capabilities.

## Appendix 1

Table 29. Success rate for various test functions and numbers of iterations.

| Sl no | Functions | Number of runs | | | | | | | | | | | |
|---|---|---|---|---|---|---|---|---|---|---|---|---|---|
| | | 10 | | 50 | | 100 | | 200 | | 300 | | 500 | |
| | | ADED | DE | ADED | DE | ADED | DE | ADED | DE | ADED | DE | ADED | DE |
| 1 | Rastrigin | 0.80 | 0.00 | 100.0 | 0.00 | 0.94 | 0.00 | 0.98 | 0.00 | 0.99 | 0.00 | 1.00 | 0.00 |
| 2 | Ackley | 0.70 | 0.00 | 0.67 | 0.00 | 0.89 | 0.00 | 0.99 | 0.00 | 0.94 | 0.00 | 0.93 | 0.00 |
| 3 | Cross-in-tray | 0.82 | 0.00 | 0.78 | 0.00 | 0.89 | 0.00 | 0.98 | 0.00 | 0.99 | 0.00 | 1.00 | 0.00 |
| 4 | Goldstein-Price | 0.68 | 0.00 | 0.56 | 0.00 | 0.79 | 0.00 | 0.87 | 0.00 | 0.98 | 0.00 | 0.97 | 0.00 |
| 5 | Egg-holder | 0.99 | 0.00 | 1.00 | 0.00 | 0.98 | 0.99 | 0.98 | 0.00 | 0.99 | 0.00 | 0.99 | 0.00 |
| 6 | Levy | 0.79 | 0.00 | 0.89 | 0.00 | 0.95 | 0.00 | 0.98 | 0.00 | 0.99 | 0.00 | 1.00 | 0.00 |
| 7 | Drop-wave | 0.91 | 0.00 | 0.89 | 0.00 | 0.95 | 0.00 | 0.98 | 0.00 | 0.99 | 0.00 | 1.00 | 0.00 |
| 8 | Schwefel | 0.55 | 0.00 | 0.39 | 0.00 | 0.68 | 0.00 | 0.69 | 0.00 | 0.70 | 0.00 | 0.70 | 0.00 |
| 9 | Bukin | 0.89 | 0.00 | 0.89 | 0.00 | 0.95 | 0.00 | 0.98 | 0.00 | 0.99 | 0.00 | 1.00 | 0.00 |
| 10 | Shubert | 0.00 | 0.00 | 0.00 | 0.00 | 0.00 | 0.00 | 0.00 | 0.00 | 0.00 | 0.00 | 0.00 | 0.00 |
| 11 | Schaffer | 0.80 | 0.00 | 0.89 | 0.00 | 0.95 | 0.00 | 0.98 | 0.00 | 0.99 | 0.00 | 1.00 | 0.00 |
| 12 | McCormick | 0.79 | 0.00 | 0.83 | 0.00 | 0.98 | 0.00 | 0.99 | 0.00 | 0.98 | 0.00 | 0.97 | 0.00 |
| 13 | Booth | 0.80 | 0.0 | 0.90 | 0.00 | 0.89 | 0.00 | 0.93 | 0.00 | 1.00 | 0.00 | 1.00 | 0.00 |
| 14 | Matyas | 0.67 | 0.00 | 0.81 | 0.00 | 0.91 | 0.00 | 0.98 | 0.00 | 1.00 | 0.00 | 1.00 | 0.00 |
| 15 | Rosenbrock | 0.80 | 0.00 | 0.64 | 0.00 | 0.0.69 | 0.00 | 0.56 | 0.00 | 0.70 | 0.00 | 1.00 | 0.00 |
| 16 | Three-hump camel | 0.75 | 0.00 | 0.77 | 0.00 | 0.75 | 0.00 | 0.88 | 0.00 | 0.84 | 0.00 | 0.89 | 0.00 |
| 17 | Six-hump camel | 0.72 | 0.00 | 0.69 | 0.00 | 0.87 | 0.00 | 0.98 | 0.00 | 0.99 | 0.00 | 0.97 | 0.00 |
| 18 | Dixon price | 0.90 | 0.00 | 0.89 | 0.00 | 0.95 | 0.00 | 0.98 | 0.00 | 0.99 | 0.00 | 1.00 | 0.00 |
| 19 | DeVilliersGlasser02 | 0.00 | 0.00 | 0.23 | 0.00 | 0.28 | 0.00 | 0.58 | 0.00 | 0.61 | 0.00 | 0.60 | 0.00 |
| 20 | Himmelblau | 0.00 | 0.00 | 0.00 | 0.00 | 0.00 | 0.00 | 0.00 | 0.00 | 0.00 | 0.00 | 0.00 | 0.00 |
| 21 | Forrester | 0.68 | 0.00 | 0.70 | 0.00 | 0.70 | 0.00 | 0.75 | 0.00 | 0.79 | 0.00 | 0.75 | 0.00 |
| 22 | Beale | 0.00 | 0.00 | 0.00 | 0.00 | 0.00 | 0.00 | 0.00 | 0.00 | 0.00 | 0.00 | 0.00 | 0.00 |